\pdfoutput=1

\documentclass[11pt,table,dvipsnames]{article}

\usepackage{acl}
\usepackage{times}
\usepackage{latexsym}
\usepackage[T1]{fontenc}

\usepackage[utf8]{inputenc}

\usepackage{microtype}
\usepackage{latexsym}
\usepackage{dsfont}
\usepackage{booktabs} 
\usepackage{subfigure}
\usepackage{graphicx}
\usepackage{bigstrut}
\usepackage{multirow}
\usepackage{floatrow}

\usepackage[normalem]{ulem}
\usepackage{amssymb}
\usepackage{amsfonts}
\usepackage{multicol}
\usepackage{tipa}
\usepackage[ruled,linesnumbered]{algorithm2e}

\usepackage{paralist}
\usepackage[switch]{lineno}
\usepackage{multirow, makecell, caption}
\usepackage{arydshln}
\usepackage{verbatim}

\usepackage{csquotes}
\usepackage{xspace}
\usepackage{paralist}
\usepackage{mdwlist}
\usepackage{subfigure}
\usepackage{array}
\usepackage{enumitem}
\usepackage{bm}
\usepackage{colortbl}
\usepackage{dsfont}
\usepackage{url}
\usepackage{graphicx}
\usepackage{tabularx}
\usepackage{amsmath}
\usepackage{bbm}
\usepackage{tikz}
\usepackage{pgfplots}
\usepackage{pifont}
\newcommand{\dataset}{\texttt{CSK-PN}\xspace}

\newcommand{\ie}{\textit{i.e.}\xspace}
\newcommand{\eg}{\textit{e.g.}\xspace}
\newcommand{\commongen}{\textsc{CommonGen}\xspace}

\newcommand{\pac}{\texttt{TP}\xspace}
\newcommand{\nac}{\texttt{TN}\xspace}
\newcommand{\aac}{\texttt{Acc}\xspace}

\makeatletter
\def\adl@drawiv#1#2#3{%
        \hskip.5\tabcolsep
        \xleaders#3{#2.5\@tempdimb #1{1}#2.5\@tempdimb}%
                #2\z@ plus1fil minus1fil\relax
        \hskip.5\tabcolsep}
\newcommand{\cdashlinelr}[1]{%
  \noalign{\vskip\aboverulesep
           \global\let\@dashdrawstore\adl@draw
           \global\let\adl@draw\adl@drawiv}
  \cdashline{#1}
  \noalign{\global\let\adl@draw\@dashdrawstore
           \vskip\belowrulesep}}
\makeatother

\title{
\textit{Say What You Mean!} 
Large Language Models Speak Too Positively about Negative Commonsense Knowledge}

\author{Jiangjie Chen\textsuperscript{\rm $\spadesuit$}, 
 Wei Shi\textsuperscript{\rm $\spadesuit$}, 
 Ziquan Fu\textsuperscript{\rm $\heartsuit$}\thanks{~~Work done while at Brain Technologies, Inc.}, 
 Sijie Cheng\textsuperscript{\rm $\spadesuit$}, 
 Lei Li\textsuperscript{\rm $\clubsuit$},
 Yanghua Xiao\textsuperscript{\rm $\spadesuit\diamondsuit$}\thanks{~~Corresponding author.}\\
\textsuperscript{\rm $\spadesuit$}Shanghai Key Laboratory of Data Science, School of Computer Science, Fudan University\\
\textsuperscript{\rm $\heartsuit$}System Inc.
\textsuperscript{\rm $\clubsuit$}University of California, Santa Barbara \\
\textsuperscript{\rm $\diamondsuit$}Fudan-Aishu Cognitive Intelligence Joint Research Center\\
\texttt{\{jjchen19, sjcheng20, shawyh\}@fudan.edu.cn}\\
\texttt{wshi22@m.fudan.edu.cn, frank@system.com, leili@cs.ucsb.edu}
}

\begin{document}

\maketitle

\begin{abstract}
Large language models (LLMs) have been widely studied for their ability to store and utilize positive knowledge. 
However, negative knowledge, such as ``\textit{lions don't live in the ocean}'', is also ubiquitous in the world but rarely mentioned explicitly in the text.
\textit{What do LLMs know about negative knowledge?}
This work examines the ability of LLMs to negative commonsense knowledge.
We design a constrained keywords-to-sentence generation task (CG) and a Boolean question-answering task (QA) to probe LLMs.
Our experiments reveal that LLMs frequently fail to generate valid sentences grounded in negative commonsense knowledge, yet they can correctly answer polar yes-or-no questions.
We term this phenomenon the \textit{belief conflict} of LLMs.
Our further analysis shows that statistical shortcuts and negation reporting bias from language modeling pre-training cause this conflict.\footnote{Resources of this paper are available at \url{https://github.com/jiangjiechen/uncommongen}.}
\end{abstract}

\section{Introduction}
\label{sec:intro}

Most of the world knowledge exists in a positive and affirmative form~\cite{molnar2000truthmakers,barker2012being,vrandevcic2014wikidata,Speer_Chin_Havasi_2017}.
As a result, large language models (LLMs) pre-trained on a colossal amount of texts, such as GPT-3~\cite{NEURIPS2020_1457c0d6,ouyang2022training} and PaLM~\cite{chowdhery2022palm}, have demonstrated their remarkable abilities for storing and utilizing positive knowledge in downstream tasks.
In contrast, negative knowledge, such as the commonsense statement that ``\textit{lions do not live in the ocean}'', is rarely mentioned in the textual world~\cite{hossain-etal-2022-analysis}.\footnote{\citet{hossain-etal-2022-analysis} report that sentences with negation hold up to 14.5\% in the CommonsenseQA dataset~\cite{talmor-etal-2019-commonsenseqa}, 8.7\% in QNLI~\cite{rajpurkar-etal-2016-squad}, and 22.6-29.9\% in general-purposed texts.}
Such negative knowledge also exists in the real world, and is important for cognitive skills such as knowing \textit{what is not true} or \textit{what not to think}~\cite{1020071ar,Minsky1997NegativeE,barker2012being}.
Therefore, we ask this question: \textit{Do LLMs (such as GPT-3 models) acquire such implicit negative knowledge through extensive language modeling pre-training?}

\begin{figure}[t!]
    \centering
    \includegraphics[width=\linewidth]{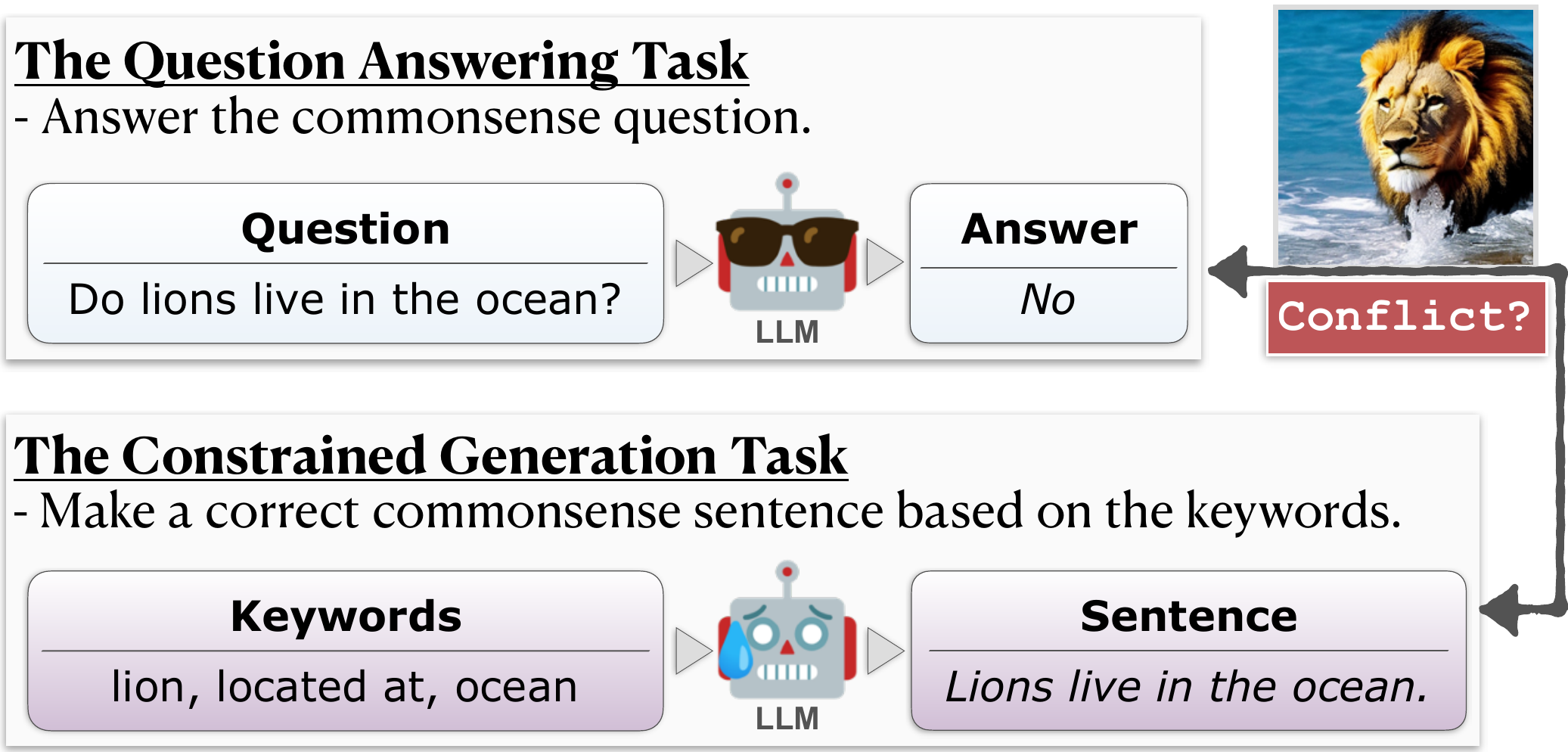}
    \caption{An example of the probing tasks studied in this paper. For the same negative commonsense knowledge <\textit{lion, located at, ocean}> which is false, we find LLMs often fail to generate texts grounded in such negative knowledge while knowing its validity according to question answering.}
    \label{fig:front}
\end{figure}

One important way of probing LLMs, which are mostly generative models, is checking whether the generated texts are knowledge-grounded.
This is because the generation of texts is a direct manifestation of a model's internal beliefs towards world knowledge~\cite{kassner-etal-2021-beliefbank,sumers2021extending,tafjord2022entailer}.\footnote{Our definition of belief is derived from \citet{kassner-etal-2021-beliefbank}, which is the assignment of a truth value to a proposition. 
In our study, the context for the proposition is the world knowledge that models learned.
Therefore, we define a model's belief about such knowledge as its prediction about the truth value of a certain piece of world knowledge.}
Knowledge-grounded text generation has been a focus of NLP research~\cite{yu2022survey}.
For example, the \commongen benchmark~\cite{lin-etal-2020-commongen} evaluates generative commonsense reasoning that organizes concepts as keyword input and generates a sentence grounded in commonsense knowledge.
However, previous work does not consider negative knowledge, nor do they probe the consistency between what models know and what they generate.
Another line of work on probing~\cite{petroni-etal-2019-language,ettinger-2020-bert,kassner-schutze-2020-negated,cao-etal-2021-knowledgeable} is conducted through the mask-infilling task.
However, this task mainly evaluates bidirectional models~\cite{devlin-etal-2019-bert}, and is not natural for unidirectional LLMs.
Also, this task suffers from the \textit{open-world problem} in evaluation, \ie, there could be multiple valid answers to fill the mask.
This is vital for evaluating negative knowledge, which has an infinite answer space, \eg, lions don't live in the \textit{sky, water, desk, car}, etc.


In this study, we investigate the belief of LLMs about negative commonsense knowledge through the lens of \textit{text generation}. 
Since LLMs have become a foundational service~\cite{bommasani2021opportunities} and cannot be easily trained, we apply in-context learning~\cite{NEURIPS2020_1457c0d6} for the probing tasks, which is tuning-free.
We design a Constrained Sentence Generation (CG) probing task, following \citet{lin-etal-2020-commongen}, where the model must generate a knowledge-grounded sentence based on a given triple <$s, r, o$>.
For example, given a triple ``<\textit{lion}, \textit{located at}, \textit{ocean}>'', a model should generate ``\textit{lions \underline{do not} live in the ocean}''.
This task is rather simple and clear.
The output sentence basically contains the same information as the input keywords.
Thus, the generated texts are easy to evaluate according to the appearance of negation.
We also add a Boolean Question Answering (QA) task that asks LLMs whether a knowledge triple is valid, which shows their beliefs about this piece of knowledge.
An example is given in Figure~\ref{fig:front}.


In our experiments, we find that LLMs of different sizes and shapes often produce hallucinated claims of negative knowledge, even if they answer yes-or-no questions about it correctly.
We term this phenomenon the \textit{belief conflict}, \ie, actions (generating texts with it) conflict with its belief (answering question about it).
Hallucinated generation of negative knowledge is seen in both our probing tasks and downstream tasks, such as explanation generation~\cite{chen-etal-2022-e,jung2022maieutic}, where negative knowledge plays an important role in the argumentation of refutation.
Further analysis shows that this problem stems from the statistical shortcuts and reporting bias of negation during pre-training.
Moreover, such implicit biases can be alleviated through explicit reasoning with Chain-of-Thought prompting~\cite{wei2022chain}, such as syllogistic deduction and related fact comparison.

The main contributions of this paper are summarized as follows:
\begin{inparaenum}[\it 1)]
    \item We are the first to investigate LLMs' belief about negative knowledge in the commonsense domain, which may shed light on a previously unstudied aspect of LLMs' abilities.
    \item We propose to probe generative LLMs through constrained sentence generation, which is effective for evaluating generated texts grounded in positive and negative knowledge.
    \item Through extensive experiments, we identify and analyze LLMs' \textit{belief conflict} phenomenon on negative commonsense knowledge, and provide insights on the causes and solutions of such problems.
\end{inparaenum}

\section{Related Work}
\label{sec:related}

\paragraph{Negative Knowledge}
Negative knowledge refers to information that describes what is not true, what cannot be done, or what does not exist, while everything that exists is positive~\cite{molnar2000truthmakers,barker2012being}. 
It plays an important role in the human reasoning process, because to think effectively, we need to know what ``not to think''~\cite{Minsky1997NegativeE}.
Current research of negative knowledge in NLP mainly focuses on developing negative knowledge bases that store relational negative commonsense knowledge \cite{10.1145/3442442.3452339,safavi-etal-2021-negater,10.1145/3511808.3557484} and utilizing negative knowledge within arguments or explanations to refute a candidate~\cite{NEURIPS2018_4c7a167b,aggarwal-etal-2021-explanations,chen-etal-2022-e}.
This paper is based on these resources to probe the belief of LLMs about the relations of everyday concepts that are not true.

\paragraph{Understanding Negation in Texts}

The manifestation of negative knowledge in texts is the phenomenon of negation~\cite{sep-negation}, which is difficult for pre-trained LMs to understand, \eg, filling ``\textit{birds cannot} \texttt{[MASK]}'' with ``\textit{fly}''~\cite{kassner-schutze-2020-negated}.
Negation has been shown to be spuriously correlated with negative or contradictory labels due to the data distribution~\cite{gururangan-etal-2018-annotation,ettinger-2020-bert,lai-etal-2021-machine,branco-etal-2021-shortcutted,tian2022debiasing}, raising doubts about the performance of previous models.
Furthermore, LMs may ignore the existence of negative words when understanding texts~\cite{kassner-schutze-2020-negated} or processing prompts~\cite{jang2022can}, which can be alleviated with unlikelihood training objective~\cite{Welleck2020Neural} during training~\cite{hosseini-etal-2021-understanding} or specifying pragmatic contexts~\cite{gubelmann-handschuh-2022-context}. 
While most current research focuses on NLU, this work fills in a gap in the investigation of the negation phenomenon in the context of text generation.

\paragraph{Knowledge-Grounded Language Models}

A major goal of NLP has been to ground LMs in world knowledge, such as factual knowledge~\cite{vrandevcic2014wikidata} and commonsense knowledge~\cite{Speer_Chin_Havasi_2017}.
A line of work~\cite{petroni-etal-2019-language,kassner-schutze-2020-negated,cao-etal-2021-knowledgeable} directly probes the knowledge implicitly learned by LMs through mask-infilling.
However, such a probing paradigm only works for contextual LMs such as BERT~\cite{devlin-etal-2019-bert}, leaving generative ones, especially modern LLMs, understudied.
Another line of work focuses on making LM-generated sentences grounded in knowledge~\cite{petroni2020how,Liu_Wan_He_Peng_Yu_2021}.
\citet{lin-etal-2020-commongen} designed a constrained text generation task, \commongen, which asks a model to generate a sentence given a set of concepts, testing the generative commonsense reasoning of LMs.
However, these studies do not investigate text generation grounded in negative knowledge, which is the focus of this work.

\paragraph{In-Context Learning}

In-context learning (ICL; \citealp{NEURIPS2020_1457c0d6}) has become a prevailing paradigm for deploying LLMs (\eg, the GPT-3 family~\citealp{NEURIPS2020_1457c0d6,chen2021evaluating,ouyang2022training}) for downstream tasks.
Through ICL, LLMs can solve tasks directly based on input-output examples without parameter updates~\cite{min-etal-2022-metaicl,rubin-etal-2022-learning}. 
Furthermore, recent work~\cite{wei2022chain,wang2022self} reveals that the ceiling performance determined by the scaling law can be beaten with ICL by generating immediate rationales, \ie, the Chain of Thought (CoT) prompting.
Since LLMs are becoming a foundational service that do not need fine-tuning, our probing on LLMs are based on ICL.


\section{Probing Protocol}
\label{sec:evaluation}

In this section, we set up an evaluation protocol to understand what LLMs know about (negative) commonsense knowledge of everyday concepts.

\subsection{The \dataset Dataset}

We limit the scope of the knowledge probed to relational knowledge between commonsense concepts, \ie, \textit{relational knowledge triples},  which exist widely in knowledge graphs and are commonly studied by the community~\cite{auer2007dbpedia,vrandevcic2014wikidata,Speer_Chin_Havasi_2017}.
Given a triplet in the form of <$s,r,o$> with a subject concept $s$, a relation $r$ and an object concept $o$, we define a negative fact as $\neg r(s, o)$ if the truth value of $r(s, o)$ is \texttt{False} according to commonsense knowledge, and a (positive) fact if otherwise.

\paragraph{Dataset Statistics}
\label{appendix:dataset}
\begin{figure}[t]
    \centering
    \includegraphics[width=\linewidth]{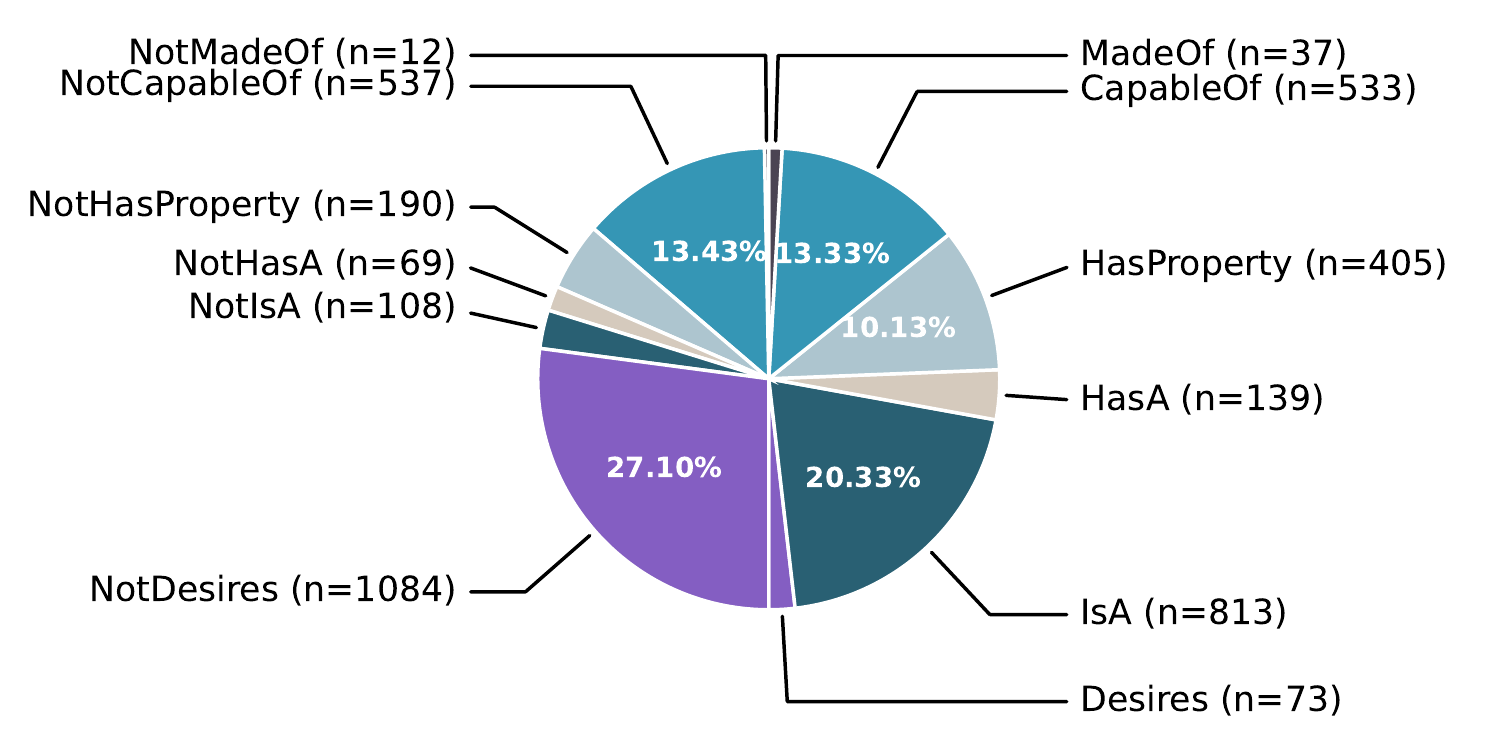}
    \caption{The configuration of the \dataset dataset.}
    \label{fig:dataset}
\end{figure}

We build the probing dataset (denoted as \dataset) based on the knowledge triples filtered by \citet{safavi-etal-2021-negater}, which are the challenging ones sourced from ConceptNet~\cite{Speer_Chin_Havasi_2017}.
We also remove invalid triples with pronouns, negation, and adjectives as subjects or objects.
The final dataset contains a total of 4,000 triples with six pairs of positive or negative relations (\eg, \textsc{IsA} and \textsc{NotIsA}), and the positive and negative splits have the same size (1:1).
Detailed information of \dataset is shown in Figure~\ref{fig:dataset}.

\subsection{Probing Task Formulation}

The most commonly used probing task for understanding whether LMs have certain types of knowledge is mask-infilling~\cite{devlin-etal-2019-bert,petroni2020how,kassner-schutze-2020-negated}.
However, this task is not suitable for generative LMs, as the mask must exist at the end of a sentence.

We argue that LLMs, which are mainly autoregressive text generation models~\cite{radford2019language,NEURIPS2020_1457c0d6,ouyang2022training,scao2022bloom}, should be investigated by \textit{text generation} with text decoding from a large sentence space.
Therefore, we propose to use \textit{Constrained Sentence Generation} (CG) as the primary task to investigate LLMs, coupled with \textit{Boolean Question Answering} (QA) for comparison, which is a common approach to probing the belief of models~\cite{tafjord2022entailer,richardson2022breakpoint}.

\paragraph{Task 1: Boolean Question Answering (QA)} 

The Boolean QA task requires LLMs to express its belief about a fact by answering a yes-or-no question.
We first transform every triplet <$s, r, o$> into a yes or no question $q$, where we remove the negation in $r$ for negative facts.
For example, a prompt goes like this:
\begin{quote}
\small{
    \textit{Answer commonsense questions with yes or no:}\\
    {\color{gray}(\textit{Examples for in-context learning})}\\
    \textbf{Question}: do lions live in the ocean?\\
    \textbf{Answer}: \underline{no}
    }
\end{quote}
where \underline{underlined texts} are completed by LLMs.
To generate the questions, we adopt InstructGPT using in-context learning ($\mathsection$\ref{sec:icl}).
The questions are 94\% valid according to a manual inspection of 50 random cases.\footnote{Bad cases are mostly due to the quality of the triples, \eg, <\textit{swim, has property, full of water}>: \textit{is swimming full of water?}}

\paragraph{Task 2: Constrained Sentence Generation (CG)}

Generating texts is a direct manifestation of a model's belief.
However, evaluating generated texts is notoriously difficult in NLP, especially without references.
Therefore, we design a \textit{keyword-to-sentence} task to make the probing more controllable, which is similar to \commongen~\cite{lin-etal-2020-commongen}.
Given a triple <$s, r, o$>, models need to generate sentences grounded in (negative) knowledge, \ie, add negation cues (\eg, \textit{not}, \textit{unable}) in the sentence if necessary, \eg,
\begin{quote}
    \small{
    \textit{Write a short and factual sentence according to commonsense based on the keywords:}\\
    {\color{gray}(\textit{Examples for in-context learning})}\\
    \textbf{Keywords}: lion, located at, ocean\\
    \textbf{Sentence}: lions don't live in the ocean.
    }
\end{quote}
We remove the \textsc{Not} prefix from the negated relations.
Note that we allow the paraphrasing of the input keywords, making it a \textit{soft}-constrained sentence generation task.

\subsection{Evaluation Metrics}

\paragraph{Metric for QA}

The QA task can be easily evaluated by checking the generated token \textit{yes} and \textit{no} (cased and uncased).
We define \pac and \nac as the accuracy on the positive and negative splits in \dataset, and \aac as the accuracy on the whole dataset (\ie, $\aac = (\pac + \nac) / 2$, since the positive and negative splits have equal size).
For rare scenarios ($<1\%$) that LLMs do not generate a \texttt{yes} or \texttt{no} token, we compare the conditional probability of these two tokens.

\paragraph{Metric for CG}

Due to the controlled task setting, which essentially forces LLMs to decide whether and how to add a negation cue during decoding, the CG task can be efficiently evaluated by detecting the existence of \textit{negation cues} (\eg, not, unable, etc.) in the generations.
Following the QA task, we also use \pac and \nac as accuracy metrics.
To implement this metric, we first use keywords-based matching for negation cues, followed by a RoBERTa model~\cite{liu2019roberta} as a \textit{token classifier} looking for unmatched negation cues.\footnote{The model is trained on the \textsc{CondaQA} dataset~\cite{ravichander2022condaqa}, which has 14,182 QA pairs with more than 200 unique negation cues.}
This metric produces 1 or 0 based on the finding of negation cues in a sentence.
After manual inspection of 200 cases, we find that this metric is correct 97\% of the time, which is reliable for evaluating such a constrained probing task.
Errors are mostly due to double negations and ambiguous negative cues (\eg, \textit{less}, \textit{opposite}, etc.), which are quite rare.

\paragraph{\textit{Can we trust negation detection as the metric to evaluate CG?}}

We manually evaluate the factuality of generated texts based on commonsense knowledge and see whether the CG metric (detection of negation) correlates well with humans in this task. 
Note that only the sentences that make common sense and adhere to the keywords constraints are accepted as true during manual annotation.
After examining 100 cases, we find that the agreement between human judgment and this metric achieves 95\%.
This is predictable, since this task is rather easy and constrained, yet LLMs do not solve it well, especially not very consistent with the QA task.
Errors made by the metric are mostly because 
\begin{inparaenum}[\it 1)]
    \item generated sentences use uncertain adverbs to modify the sentences, \eg, \textit{may}, \textit{some}, etc.;
    \item noisy triples in the dataset.
\end{inparaenum}
Overall, we think this metric is trustworthy and evaluates this task far better than most popular text generation metrics.

\section{\textit{Do LLMs have negative commonsense knowledge?}}
\label{sec:results}

In this section, we use \dataset to investigate LLMs' belief about negative commonsense knowledge.
More importantly, \textit{can LLMs generate texts grounded in negative commonsense knowledge?}

\subsection{Probing LLMs with In-Context Learning}
\label{sec:icl}

To execute the probing tasks without fine-tuning, we exploit the few-shot in-context learning (\citealp{NEURIPS2020_1457c0d6}) ability of LLMs.
We manually write 32 examples, with 16 examples for positive knowledge (denoted as $E^+$) and 16 for negative knowledge ($E^-$).\footnote{Examples can be found in Appendix~\ref{appendix:demo4probing}}
In the experiments, we randomly sample a total number of $k$ examples from $E^+$ and $E^-$, where $|E^+| = |E^-|$ if not specified.\footnote{Example prompts for two tasks are in Appendix~\ref{appendix:icl_ex}.}

\paragraph{Choices of LLMs}

\newcolumntype{a}{>{\columncolor{BlueGreen!10}}c}
\newcolumntype{b}{>{\columncolor{Blue!10}}c}
\newcolumntype{d}{>{\columncolor{Purple!10}}c}

\begin{table}[t]
    \small
    \centering
    \begin{tabular}{ccaaabbbd}
    \toprule
        \multirow{2}{*}{\textbf{Model}} & \multirow{2}{*}{$k$} & \multicolumn{3}{c}{\textbf{Perf. on QA}} & \multicolumn{3}{c}{\textbf{Perf. on CG}} & \multicolumn{1}{c}{\multirow{2}{*}{\textbf{Cns.}}} \\
        \cmidrule(lr){3-5}
        \cmidrule(lr){6-8}
        &  & \multicolumn{1}{c}{\pac} & \multicolumn{1}{c}{\nac} & \multicolumn{1}{c}{\aac} & \multicolumn{1}{c}{\pac} & \multicolumn{1}{c}{\nac} &\multicolumn{1}{c}{\aac} & \multicolumn{1}{c}{}\\
    \midrule
         \multirow{2}{*}{\makecell[c]{Flan-T5\\(3B)}}  
         & 2 & 79.1 & 84.0 & 81.5 & 96.5 & 19.4 & 57.9 & 56.2 \\
         &  10 & 82.7 & 80.2 & 81.4 &  96.9 & 19.8 &  58.4 & 59.7 \\
         \cdashlinelr{1-9}
         \multirow{2}{*}{\makecell[c]{Flan-T5\\(11B)}} 
        & 2 & 84.1 & 81.0 & 82.6 & \underline{97.5}  & 15.9 & 56.7 & 57.7\\
         &  10 & 85.4 & 80.8 & 83.1 &  \textbf{97.6} & 28.2 & 62.9 & 65.9\\
        \midrule
        \multirow{2}{*}{\makecell[c]{GPT-3}}
         & 2 &  76.0 &  58.9 & 67.5  & 83.9  & 28.4 & 56.1  & 54.4\\
        & 10 & 74.7  & 66.9  & 70.8  &  30.9 & \textbf{79.8} &  55.3 & 53.7\\
        \midrule
        \multirow{2}{*}{\makecell[c]{Codex$_\texttt{002}$}} 
         &  2 & \textbf{89.2}  &  81.7 & \textbf{85.4}  & 96.6  & 38.0 & 67.3 & 70.1 \\
        &  10 &  88.1 & 81.8  & \underline{84.9} & 93.2  & 68.8 & 81.0  & \underline{84.5}\\
        \midrule
        \multirow{2}{*}{\makecell[c]{Instruct-\\GPT$_\texttt{001}^\texttt{curie}$}}
         &  2 & 85.2& 51.1 & 68.2  &  90.1 & 21.9 & 56.0 & 67.3 \\
        & 10 & 70.0 & 65.8 & 67.9 & 71.5 & 40.8 & 56.1 & 58.2 \\
         \cdashlinelr{1-9}
        \multirow{2}{*}{\makecell[c]{Instruct-\\GPT$_\texttt{001}$}} 
         &  2 & 78.1 & 83.6  & 80.9  &  94.9 & 25.0 & 60.0 & 57.7 \\
        & 10 & 79.5 & 81.6 & 80.6 & 79.2  & 55.4 & 67.3 & 68.2 \\
         \cdashlinelr{1-9}
        \multirow{2}{*}{\makecell[c]{Instruct-\\GPT$_\texttt{002}$}} 
         & 2 & 81.7 & \textbf{86.1} & 83.9 & 92.9  & 48.7 &  72.1 & 71.2\\
         &  10 & 84.1 & \underline{84.7} & 84.4 & 88.9  & 61.4 & 75.1 & 77.5\\
         \cdashlinelr{1-9}
        \multirow{2}{*}{\makecell[c]{Instruct-\\GPT$_\texttt{003}$}} 
         & 2 & 87.9 & {81.3} & 84.6 & 95.1  & 58.1 &  76.6 & 80.5\\
         &  10 & \underline{89.0} & {79.5} & 84.2 & 91.1  & 73.6 & \underline{82.3} & \textbf{87.9} \\
         \midrule
         \multirow{2}{*}{ChatGPT} 
         & 2 & 82.9 & 82.0 & 82.4 & 89.8 & 69.8 & 79.8 & 79.2 \\
         & 10 & 81.5 & 85.7 & 83.6 & 90.4 & \underline{78.4} & \textbf{84.4} & 84.1 \\
    \bottomrule
    \end{tabular}
    \caption{Main results of different LLMs, which are obtained with $k$ examples ($|E^+|=|E^-|$). \textbf{Cns.} denotes the consistency between QA and CG.
    The best results are \textbf{bolded} and the second best are \underline{underlined}.
    }
    \label{tab:bc_struggle}
\end{table}

We use LLMs that can do in-context learning, so that models stay fixed during probing.
We choose Flan-T5~\cite{chung2022scaling}, GPT-3~(175B, \texttt{davinci}; \citealp{NEURIPS2020_1457c0d6}) and GPT-3.5 series, \eg Codex ($\ge$175B, \texttt{code-davinci-002}; \citealp{chen2021evaluating}) and InstructGPT~\cite{ouyang2022training}: all are capable of in-context learning.
Flan-T5 is an encoder-decoder LLM with instruction tuning based on T5~\cite{raffel2020exploring}.
Codex extends GPT-3 through code training and instruction fine-tuning, and InstructGPT extends Codex through further tuning of the instructions.
In our experiments, we mainly explore GPT-3.5 models.
We use the 6.7B variant of InstructGPT (\texttt{text-curie-001}) and the $\ge$175B variants, \ie, \texttt{text-davinci-001} (tuned on instructions), 
\texttt{text-davinci-002} (tuned on code and instructions), and \texttt{text-davinci-003} (further tuned with reinforcement learning with human feedback, RLHF).\footnote{\url{https://beta.openai.com/docs/model-index-for-researchers}}
For deterministic predictions, all models use greedy decoding (temperature as $0.0$)\footnote{We find our findings in the experiments are consistent for different temperatures, according to Appendix~\ref{appendix:temperature}.}.
We use InstructGPT$_\texttt{002}$ as the default LLM for experiments due to its powerful capability and the fact that it has been extensively researched and applied as of the time of writing this paper.
We also include the recent ChatGPT~\cite{openai2022chatgpt}, which is built upon InstructGPT and trained with dialogue data and RLHF.

\subsection{The Belief Conflict}
\label{sec:beliefconflict}

We report the results of the probing tasks in Table~\ref{tab:bc_struggle} for LLMs with 2- and 10-shot in-context learning.
Based on the results, we discover a clear conflict of LLMs, that LLMs behave inconsistently in QA and CG tasks on negative commonsense knowledge, which we term \textit{belief conflict}.
Such conflict manifests itself in two ways: the gap between \pac and \nac on the CG task, and the gap of \nac between the QA and CG tasks.
In general, belief conflicts exist across LLMs of various sizes and structures.
Ablated results per relation is presented in Appendix~\ref{appendix:relation}.

When specifically asked, LLMs can distinguish between positive and negative commonsense knowledge, as evidenced by stable and balanced scores for positive and negative splits in the QA task.
For CG, LLMs seem to accurately generate sentences grounded in positive knowledge according to \pac.
However, they perform poorly in negative knowledge, even for the best-performing LLMs, \ie, Codex$_\texttt{002}$, InstructGPT$_\texttt{002,003}$, as shown by the lower bars of the CG on the negative split.\footnote{The only exception is GPT-3 (\texttt{davinci}).
It scores poorly on the positive split with 10-shot learning, with \nac exceeding \pac.
This happens when $k\ge4$, while its 6.7B variant (\texttt{curie}) behaves consistently with others.
Detailed results for GPT-3 are in Appendix~\ref{appendix:gpt3}.}
Also, the inconsistency between QA and CG reflects this conflict, as the content generated by a trustworthy AI system should consistent and faithful to what it believes.
We present a case study and error analysis in Appendix~\ref{appendix:case_study}.

Among these LLMs, InstructGPT$_\texttt{003}$ and ChatGPT achieve much better results than others.
We assume that such improvements are probably a result of training LLMs with human feedback (\eg, RLHF) based on the disclosed differences between them by OpenAI. 
Another evidence is that the recent ChatGPT also expresses great capabilities of generating negative knowledge, even better than InstructGPT$_\texttt{003}$ in this regard.
We hypothesize that this is because negative knowledge and rebuttal statements are frequently used in human feedback to steer the model, \eg, admitting errors or instructing the model not to do something.
To validate this claim, future work could conduct more rigorous comparisons on public available LLMs, which would be an interesting research problem to trace certain abilities of LLMs to a specific period of training.

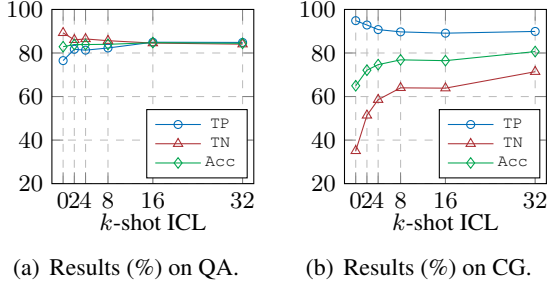
\begin{figure}[t]
    \centering
	\subfigure[Results (\%) on QA.] {
    \label{fig:abl_size_qa} 
		\pgfplotsset{width=0.345\linewidth,height=0.3\linewidth,compat=1.5,scale only axis}
\footnotesize
\begin{tikzpicture}
\begin{axis}[
    xlabel={$k$-shot ICL},
    xmin=-2, xmax=34,
    ymin=20, ymax=100,
    xtick={0, 2, 4, 8, 16, 32},
    legend pos=south east,
    ymajorgrids=true,
    xmajorgrids=true,
    grid style=dashed,
    x label style={at={(axis description cs:0.5,-0.125)},anchor=north},
    legend style={nodes={scale=0.7, transform shape},font=\footnotesize}
]
\addplot[
    color=NavyBlue,
    mark=o,
    mark size=1.5pt,
    error bars/.cd,
    y dir=both, y explicit
    ]
    coordinates {
    (0, 76.50)
    (2, 81.70)
    (4, 81.30)
    (8, 82.30)
    (16, 84.95)
    (32, 84.85)
    };
    \addlegendentry{\pac}
\addplot[
    color=Maroon,
    mark=triangle,
    mark size=2pt,
    error bars/.cd,
    y dir=both, y explicit
    ]
    coordinates {
    (0, 89.25)
    (2, 86.05)
    (4, 86.45)
    (8, 85.65)
    (16, 84.55)
    (32, 84.05)
    };
    \addlegendentry{\nac}
\addplot[
    color=Green,
    mark=diamond,
    mark size=2pt,
    error bars/.cd,
    y dir=both, y explicit
    ]
    coordinates {
    (0, 82.88)
    (2, 83.88)
    (4, 83.88)
    (8, 83.98)
    (16, 84.75)
    (32, 84.45)
    };
    \addlegendentry{\aac}   
\end{axis}
\end{tikzpicture}}
	\subfigure[Results (\%) on CG.] { 
	\label{fig:abl_size_cg} 
		\pgfplotsset{width=0.345\linewidth,height=0.3\linewidth,compat=1.5,scale only axis}
\footnotesize
\begin{tikzpicture}
\begin{axis}[
    xlabel={$k$-shot ICL},
    xmin=-2, xmax=34,
    ymin=20, ymax=100,
    xtick={0, 2, 4, 8, 16, 32},
    legend pos=south east,
    ymajorgrids=true,
    xmajorgrids=true,
    grid style=dashed,
    x label style={at={(axis description cs:0.5,-0.125)},anchor=north},
    legend style={nodes={scale=0.7, transform shape},font=\footnotesize}
]
\addplot[
    color=NavyBlue,
    mark=o,
    mark size=1.5pt,
    error bars/.cd,
    y dir=both, y explicit
    ]
    coordinates {
    (0, 94.85)
    (2, 92.85)
    (4, 90.75)
    (8, 89.7)
    (16, 89.1)
    (32, 89.9)
    };
    \addlegendentry{\pac}
\addplot[
    color=Maroon,
    mark=triangle,
    mark size=2pt,
    error bars/.cd,
    y dir=both, y explicit
    ]
    coordinates {
    (0, 35.05)
    (2, 51.35)
    (4, 58.55)
    (8, 64)
    (16, 63.85)
    (32, 71.4)
    };
    \addlegendentry{\nac}
\addplot[
    color=Green,
    mark=diamond,
    mark size=2pt,
    error bars/.cd,
    y dir=both, y explicit
    ]
    coordinates {
    (0, 64.95)
    (2, 72.1)
    (4, 74.65)
    (8, 76.85)
    (16, 76.48)
    (32, 80.65)
    };
    \addlegendentry{\aac}
\end{axis}
\end{tikzpicture}}
    \caption{Performance change for InstructGPT$_\texttt{002}$ on both tasks as the number of example ($k$) increases.}
    \label{fig:size}
\end{figure}

\paragraph{Sensitivity to the Number of In-Context Examples}
\label{appendix:size}

To find whether adding more examples helps solve the probing tasks, we increase the in-context examples from 0 to 32.
Figure~\ref{fig:abl_size_qa} shows a consistent finding with previous results, that LLMs are so good at answering yes or no questions that the number of examples does not affect much of the QA performance.
Figure~\ref{fig:abl_size_cg} shows that, adding more examples helps generate both positive and negative commonsense knowledge.
However, the gap between \pac and \nac in the CG task still exists.

\section{Analysis on the Belief Conflict}
\label{sec:analysis}

\subsection{\textit{Could keywords as task input hinder the manifestation of LLMs' belief?}}

The task input difference for CG and QA leads to a concern that LMs may find it easier to understand natural questions (QA) than keywords (CG); hence, the belief conflict.
In response to this concern, we change the input of the two tasks.
For example, the keywords-to-answer task takes the form as:
\begin{quote}
\small{
    \textit{Can these keywords form a truthful common sense fact? Answer with yes or no.}\\
    \textbf{Keywords}: lion, located at, ocean\\
    \textbf{Answer}: \underline{no}
    }
\end{quote}
As for the question-to-sentence task:
\begin{quote}
\small{
    \textit{Answer the question by writing a short sentence that contains correct common sense knowledge.}\\
    \textbf{Question}: do lions live in the ocean?\\ 
    \textbf{Sentence}: \underline{lions don't live in the ocean.}
    }
\end{quote}

\paragraph{Results}

\begin{figure}[t]
    \centering
	\subfigure[Results (\%) on QA.] {
	\label{fig:input_histo_qa}
\pgfplotsset{width=0.345\linewidth,height=0.3\linewidth,compat=1.5,scale only axis}
\footnotesize

\begin{tikzpicture}
    \begin{axis}[
        ybar,
        bar width=8pt,
        xtick distance=1,
        symbolic x coords={\pac, \nac, \aac },
        ymin=0, ymax=100,
        enlarge x limits=0.3,
        scaled ticks=false,
        legend pos=south east,
        xtick style={
            /pgfplots/major tick length=0pt,
        },
        legend style={nodes={scale=0.7, transform shape},font=\footnotesize}
    ]
        \addplot+ [
            color=Blue!30,
            draw=Blue,
            error bars/.cd,
            y dir=both,
            y explicit,
            error mark options={
              Blue,
              mark size=0.2pt,
              line width=4pt
            }
        ] coordinates {
            (\pac,83.73) +- (1.23,1.42)
            (\nac,85.54) +- (0.46,0.34)
            (\aac,84.63) +- (0.44,0.66)
        };

        \addplot+ [
            color=BlueGreen!30,
            draw=BlueGreen,
            error bars/.cd,
            y dir=both,
            y explicit relative,
        ] coordinates {
            (\pac,79.4)
            (\nac,79.65)
            (\aac,79.525)
        };

        \legend{
            \textit{question-to-answer},
            \textit{keywords-to-answer},
        }
    \end{axis}
\end{tikzpicture}}
	\subfigure[Results (\%) on CG.] {
    \label{fig:input_histo_cg}
\pgfplotsset{width=0.345\linewidth,height=0.3\linewidth,compat=1.5,scale only axis}
\footnotesize
\begin{tikzpicture}
    \begin{axis}[
        ybar,
        bar width=8pt,
        xtick distance=1,
        symbolic x coords={\pac, \nac, \aac },
        ymin=0, ymax=100,
        enlarge x limits=0.3,
        scaled ticks=false,
        legend pos=south east,
        xtick style={
            /pgfplots/major tick length=0pt,
        },
        legend style={nodes={scale=0.7, transform shape},font=\footnotesize}
    ]
        \addplot+ [
            color=Purple!30,
            draw=Purple,
            error bars/.cd,
            y dir=both,
            y explicit, 
            error mark options={
              Purple,
              mark size=0.2pt,
              line width=4pt
            }
        ] coordinates {
            (\pac,90.73) +- (0.68,1.02)
            (\nac,66.36) +- (4.19,2.36)
            (\aac,78.54) +- (2.21,1.69)
        };

        \addplot+ [
            color=Lavender!30,
            draw=Lavender,
            error bars/.cd,
            y dir=both,
            y explicit relative,
        ] coordinates {
            (\pac,84.65)
            (\nac,80.60)
            (\aac,82.625)
        };

        \legend{
            \textit{keywords-to-sentence},
            \textit{question-to-sentence}
        }
    \end{axis}
\end{tikzpicture}}
    \caption{Results of InstructGPT$_\texttt{002}$ when switching the task inputs between \textit{question} and \textit{keywords}, where $k=10$. 
    Columns with error bars show the ranges of the influence brought by different instruction wordings.
    }
    \label{fig:task_input}
\end{figure}
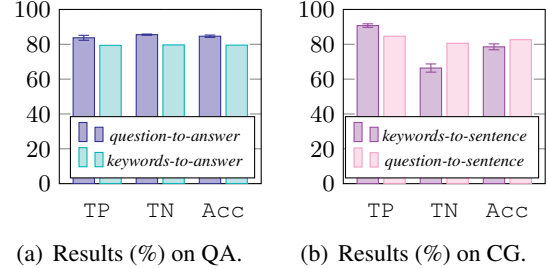

In Figure~\ref{fig:input_histo_qa}, we see a 4-point performance decrease given \textit{keywords} as input for QA, which is not significant in comparison, and the results on the positive and negative splits are as balanced as before.
{This implies that LLMs' imbalanced performance in CG is not due to the use of keywords as input.}
In Figure~\ref{fig:input_histo_cg}, CG performance is greatly improved given \textit{question} as input, approximating QA results.
Our assumption is that CG is basically transformed into QA, because the textual corpus has seen too many negated texts following a Boolean question and rephrasing it, \eg, ``\textit{...? No, lions do not live in the ocean.}''
To validate this, we provide LLMs with zero-shot question-to-sentence instructions, and check if the output sentences start with \textit{yes} or \textit{no} given an input question.
If our assumption is correct, models without examples will be biased toward QA even with a question-to-sentence instruction.
The results of models optimized for instructions show that: 84.58\% of sentences generated by InstructGPT$_\texttt{002}$ begin with yes or no, and 80.28\% for InstructGPT$_\texttt{003}$.
With 10 examples, this number drops to less than 4\%. 
Thus, {these results confirms that question-to-sentence generation degenerates to the QA task.}

As a result, {we conclude that the keyword-to-sentence (CG) is an appropriate and challenging task to probe generative LLMs.}
Employing keywords as input does not impact LLMs' grasp of the task (Figure~\ref{fig:input_histo_qa}), while using questions as input may produce shortcuts that obscure whether LLMs can generate texts of negative commonsense knowledge (Figure~\ref{fig:input_histo_cg}).
Even if we use different instruction wordings (instructions are at Appendix~\ref{appendix:icl_ex}), none escapes the belief conflict, as shown by the error bars in Figure~\ref{fig:task_input}.
Additionally, this experiment brings up the problem of how LLMs encode commonsense knowledge.
According to this experiment, commonsense knowledge seems to be stored in LLMs in the same manner as it is in the corpus.
LLMs struggle to generalize them, as evidenced by the keyword inputs for negative knowledge that do not have a statistical shortcut from pre-training.

\subsection{\textit{Will the keyword co-occurrence within corpus affect LLMs' generation?}}
\label{sec:cooccur}

LLMs are essentially statistical models.
In this experiment, we investigate the influence of \textit{word co-occurrence in the corpus} on the CG task, which is one of the most common statistical factors.
We categorize the dataset into buckets based on keywords co-occurrence on naturally existing corpora such as OMCS (706K sentences, \citealp{10.1007/3-540-36124-3_77}) and Wikipedia (1M, a subset built by \citet{gao-etal-2021-simcse}).
The co-occurrence for each triple is calculated by $\frac{\sum_{i,j} \mathtt{cooccur}(w_i, w_j)}{l_s l_o}$, where $w_i\in s, w_j\in o$, and $l_s, l_o$ denote the word count of subject $s$ and object $o$, discarding stopwords.

\begin{figure}[t]
    \centering
    \includegraphics[width=\linewidth]{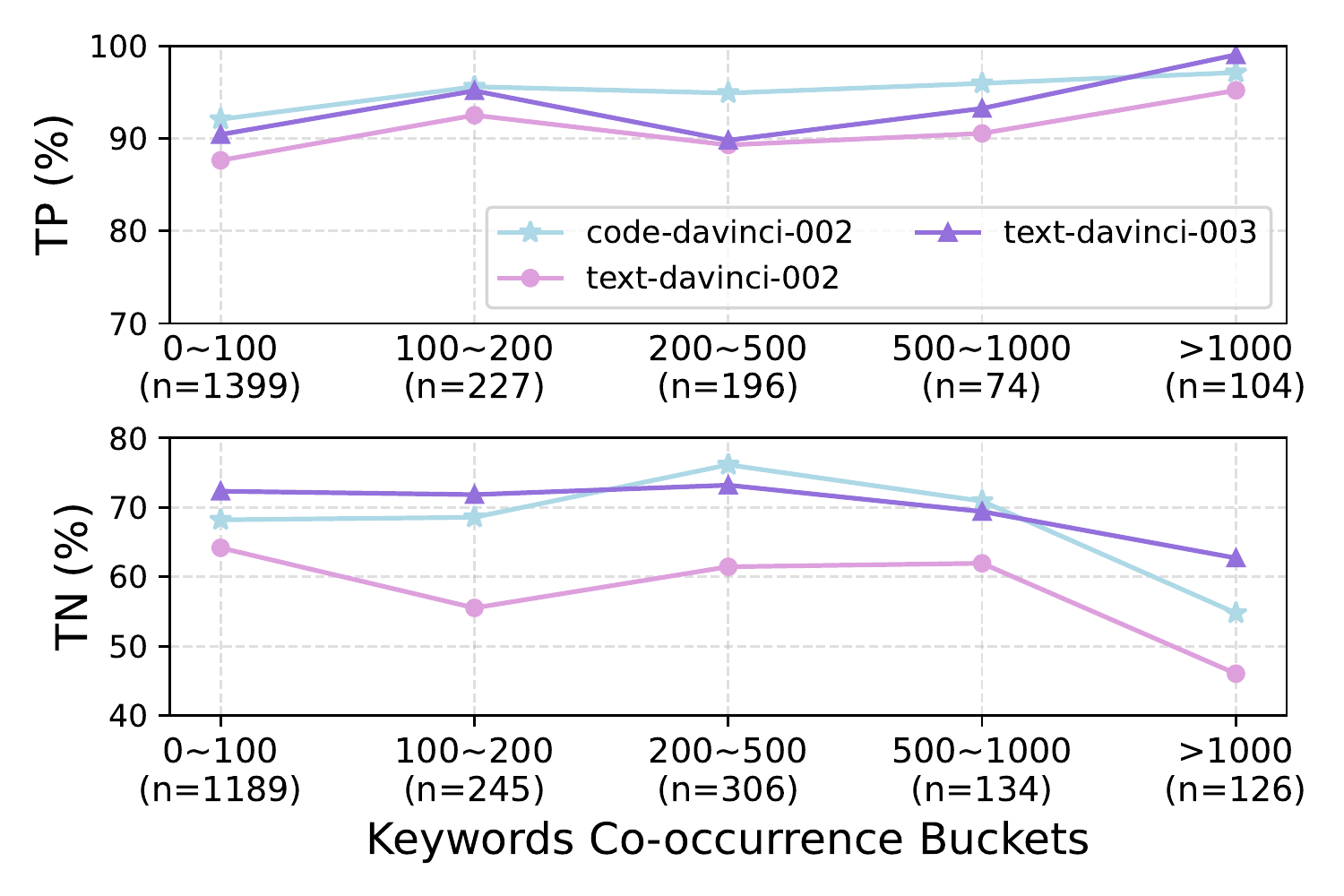}
    \caption{10-shot CG results of three best-performing LLMs on different co-occurrence buckets.
    $a\sim b$ denotes that keywords co-occurrence in a bucket ranges from $a$ to $b$.
    $n$ is the number of triples in a bucket.
    }
    \label{fig:cooccur}
\end{figure}

From Figure~\ref{fig:cooccur}, we have an interesting finding that three of the best-performing LLMs from Table~\ref{tab:bc_struggle} suffer from a performance drop at the $>1000$ bucket of the negative split (\nac), the most frequent data bucket.
In contrast, LLMs achieve the best performance this bucket on the positive split (\pac).
We conclude that the hard-to-generate negative knowledge for LLMs tend to be those in which they have seen many subjects and objects appear together.
For example, \textit{worm} and \textit{bird} usually co-occur in sentences, but models tend to generate ``\textit{worms can eat birds.}''
Such statistical shortcuts hinder the generation of negative knowledge.
This is also validated by \pac results, where LLMs find it easy to generate sentences with frequently co-occurring entities in a positive fact.

\subsection{\textit{How does the balance of positive and negative examples affect negation bias?}}

\begin{figure}[t]
    \centering
	\subfigure[Results (\%) on QA.] { 
	\label{fig:abl_ratio_qa} 
		\pgfplotsset{width=0.345\linewidth,height=0.3\linewidth,compat=1.5,scale only axis}
\small
\begin{tikzpicture}
\begin{axis}[
    xlabel={$\eta_\mathtt{QA}$ $(k=10)$},
    xmin=-0.05, xmax=1.05,
    ymin=20, ymax=100,
    xtick={0/10, 2/10, 4/10, 6/10, 8/10, 10/10},
    legend pos=south east,
    ymajorgrids=true,
    xmajorgrids=true,
    grid style=dashed,
    x label style={at={(axis description cs:0.5,-0.125)},anchor=north},
    y label style={at={(axis description cs:-0.125,0.5)},anchor=south},
    legend style={nodes={scale=0.7, transform shape},font=\footnotesize}
]
\addplot[
    color=NavyBlue,
    mark=o,
    mark size=1.5pt,
    ]
    coordinates {
    (0/10, 91.7)
    (1/10, 85.25)
    (2/10, 84.85)
    (3/10, 85.00)
    (4/10, 84.00)
    (5/10, 84.05)
    (6/10, 84.65)
    (7/10, 84.15)
    (8/10, 83.75)
    (9/10, 82.70)
    (10/10, 73.10)
    };
    \addlegendentry{\pac}
\addplot[
    color=Maroon,
    mark=triangle,
    mark size=2pt,
    ]
    coordinates {
    (0/10, 77.20)
    (1/10, 84.45)
    (2/10, 84.90)
    (3/10, 84.95)
    (4/10, 84.80)
    (5/10, 84.70)
    (6/10, 84.25)
    (7/10, 83.9)
    (8/10, 83.65)
    (9/10, 84.30)
    (10/10, 89.05)
    };
    \addlegendentry{\nac}
\addplot[
    color=Green,
    mark=diamond,
    mark size=2pt,
    ]
    coordinates {
    (0/10, 84.45)
    (1/10, 84.85)
    (2/10, 84.88)
    (3/10, 84.98)
    (4/10, 84.40)
    (5/10, 84.38)
    (6/10, 84.45)
    (7/10, 84.03)
    (8/10, 83.70)
    (9/10, 83.50)
    (10/10, 81.075)
    };
    \addlegendentry{\aac}
\end{axis}
\end{tikzpicture}}
	\subfigure[Results (\%) on CG.] { 
    \label{fig:abl_ratio_cg} 
		\pgfplotsset{width=0.345\linewidth,height=0.3\linewidth,compat=1.5,scale only axis}
\footnotesize
\begin{tikzpicture}
\begin{axis}[
    xlabel={$\eta_\mathtt{CG}$ $(k=10)$},
    xmin=-0.05, xmax=1.05,
    ymin=20, ymax=100,
    xtick={0/10, 2/10, 4/10, 6/10, 8/10, 10/10},
    legend pos=south east,
    ymajorgrids=true,
    xmajorgrids=true,
    grid style=dashed,
    x label style={at={(axis description cs:0.5,-0.125)},anchor=north},
    y label style={at={(axis description cs:-0.125,0.5)},anchor=south},
    legend style={nodes={scale=0.7, transform shape},font=\footnotesize}
]
\addplot[
    color=NavyBlue,
    mark=o,
    mark size=1.5pt,
    ]
    coordinates {
    (0/10, 97.40)
    (1/10, 94.84)
    (2/10, 93.76)
    (3/10, 92.00)
    (4/10, 90.28)
    (5/10, 88.60)
    (6/10, 87.32)
    (7/10, 85.32)
    (8/10, 82.84)
    (9/10, 77.56)
    (10/10, 67.28)
    };
    \addlegendentry{\pac}
\addplot[
    color=Maroon,
    mark=triangle,
    mark size=2pt,
    ]
    coordinates {
    (0/10, 30.16)
    (1/10, 48.24)
    (2/10, 55.88)
    (3/10, 59.56)
    (4/10, 61.20)
    (5/10, 62.60)
    (6/10, 66.00)
    (7/10, 68.84)
    (8/10, 72.00)
    (9/10, 76.96)
    (10/10, 84.16)
    };
    \addlegendentry{\nac}
    
\addplot[
    color=Green,
    mark=diamond,
    mark size=2pt,
    ]
    coordinates {
    (0/10, 63.78)
    (1/10, 71.54)
    (2/10, 74.82)
    (3/10, 75.78)
    (4/10, 75.74)
    (5/10, 75.60)
    (6/10, 76.66)
    (7/10, 77.08)
    (8/10, 77.42)
    (9/10, 77.26)
    (10/10, 75.72)
    };
    \addlegendentry{\aac}

\end{axis}
\end{tikzpicture}}
    \subfigure[Results (\%) on QA.] { 
    \label{fig:abl_ratio_qa_2} 
		\pgfplotsset{width=0.345\linewidth,height=0.3\linewidth,compat=1.5,scale only axis}
\footnotesize
\begin{tikzpicture}

\begin{axis}[
    xlabel={$|E^-|$ ($k=5+|E^-|$)},
    xmin=4.5, xmax=15.5,
    ymin=20, ymax=100,
    xtick={5,7,9,11,13,15},
    legend pos=south east,
    ymajorgrids=true,
    xmajorgrids=true,
    grid style=dashed,
    x label style={at={(axis description cs:0.5,-0.125)},anchor=north},
    y label style={at={(axis description cs:-0.125,0.5)},anchor=south},
    legend style={nodes={scale=0.7, transform shape},font=\footnotesize}
]
\addplot[
    color=NavyBlue,
    mark=o,
    mark size=1.5pt,
    ]
    coordinates {
    (5, 84.05)
    (7, 84.05)
    (9, 84.05)
    (11, 84.20)
    (13, 82.65)
    (15, 82.55)
    };
    \addlegendentry{\pac}
\addplot[
    color=Maroon,
    mark=triangle,
    mark size=2pt,
    ]
    coordinates {
    (5, 84.70)
    (7, 84.20)
    (9, 84.30)
    (11, 84.50)
    (13, 84.50)
    (15, 84.35)
    };
    \addlegendentry{\nac}
    
\addplot[
    color=Green,
    mark=diamond,
    mark size=2pt,
    error bars/.cd,
    y dir=both, y explicit
    ]
    coordinates {
    (5, 84.38)
    (7, 84.13)
    (9, 84.18)
    (11, 84.35)
    (13, 83.58)
    (15, 83.45)
    };
    \addlegendentry{\aac}
\end{axis}

\end{tikzpicture}}
    \subfigure[Results (\%) on CG.] { 
    \label{fig:abl_ratio_cg_2} 
		\pgfplotsset{width=0.345\linewidth,height=0.3\linewidth,compat=1.5,scale only axis}
\footnotesize
\begin{tikzpicture}
\begin{axis}[
    xlabel={$|E^-|$ ($k=5+|E^-|$)},
    xmin=4.5, xmax=15.5,
    ymin=20, ymax=100,
    xtick={5,7,9,11,13,15},
    legend pos=south east,
    ymajorgrids=true,
    xmajorgrids=true,
    grid style=dashed,
    x label style={at={(axis description cs:0.5,-0.125)},anchor=north},
    y label style={at={(axis description cs:-0.125,0.5)},anchor=south},
    legend style={nodes={scale=0.7, transform shape},font=\footnotesize}
]
\addplot[
    color=NavyBlue,
    mark=o,
    mark size=1.5pt,
    ]
    coordinates {
    (5, 88.6)
    (7, 89.96)
    (9, 89.)
    (11, 87.8)
    (13, 87.2)
    (15, 86.36)
    };
    \addlegendentry{\pac}
\addplot[
    color=Maroon,
    mark=triangle,
    mark size=2pt,
    ]
    coordinates {
    (5, 62.6)
    (7, 71.24)
    (9, 73.04)
    (11, 75.96)
    (13, 76.92)
    (15, 78.2)
    };
    \addlegendentry{\nac}
    
\addplot[
    color=Green,
    mark=diamond,
    mark size=2pt,
    error bars/.cd,
    y dir=both, y explicit
    ]
    coordinates {
    (5, 75.60)
    (7, 80.6)
    (9, 81.02)
    (11, 81.88)
    (13, 82.06)
    (15, 82.28)
    };
    \addlegendentry{\aac}

\end{axis}
\end{tikzpicture}}
    \caption{Results of InstructGPT$_\texttt{002}$ as the numbers of $E^+$ and $E^-$ change.
    Figure (a) and (b) increase $\eta = |E^-|/k$ while fixing $k=10$.
    Figure (c) and (d) add more $E^-$ while fixing $|E^+|=5$.
    }
    \label{fig:ratio}
\end{figure}
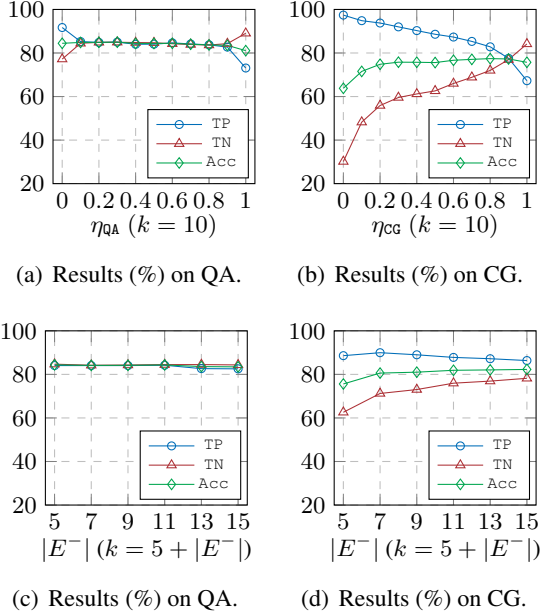

A possible answer for the difference between CG and QA is that: LMs suffer from reporting bias of negation during pre-training, while answering questions with yes or no is quite balanced in the corpora.
We validate this problem by mitigating the negation bias through adjusting the examples of positive and negative cases.
With more $E^-$s, LLMs are encouraged to generate more negations.

\paragraph{Results}

Figure~\ref{fig:abl_ratio_qa}, \ref{fig:abl_ratio_cg} adjust the ratio $\eta=\frac{|E^-|}{k}$ while fixing $k$.
Figure~\ref{fig:abl_ratio_qa} shows that InstructGPT$_\texttt{002}$ is very resilient against the example ratio in the QA task, except for extreme cases where only $E^+$s or $E^-$s are presented (\ie, $\eta \in \{0, 1\}$).
{This also demonstrates the robustness of adopting QA results as LLMs' belief.}
In Figure~\ref{fig:abl_ratio_cg}, the CG performance on the negative split is improving as $\eta$ grows.
The turning point appears somewhere near $\eta \in (0.9, 1)$ when $E^-$ takes over all the examples.
Also, \pac drops as $E^+$ becomes less.
What if we add $E^-$ without dropping $E^+$?
In Figure~\ref{fig:abl_ratio_qa_2}, \ref{fig:abl_ratio_cg_2}, we keep $E^+$ as constant ($|E^+| = 5$) and increase $|E^-|$ from $5$ to $15$.
With enough amount of $E^+$, \nac to CG continues to increase without sacrificing \pac.

Overall, {Figure~\ref{fig:ratio} presents the possibility that we can overcome the belief conflict brought about by reporting bias by increasing negated texts in the training data or in-context examples.}
However, this is not always feasible in practice.


\subsection{\textit{Do Chain-of-Thought help generate texts with negative commonsense knowledge?}}

Can the implicit reporting bias be overcome by explicit reasoning?
Recent studies~\cite{wei2022chain,wei2022emergent} discover that the Chain-of-Thought (CoT) prompting technique shows the emergent reasoning abilities of LLMs.
CoT generates intermediate steps in natural language, extending <input, output> to <input, \textit{chain-of-thought}, output>.
We adopt two instances of CoT: deductive reasoning and fact comparison, whose examples are manually written, which are in Appendix~\ref{appendix:demo4probing}.

\paragraph{Deductive Reasoning Prompting}

We instantiate CoT with deductive argumentation in the form of \textit{syllogism} (two premises and one conclusion).
The prompt is extended into <input, ``\textit{Let's think step by step}: ...'', output> with intermediate steps.
A natural way to identify a negative proposition is deductive reasoning with \textit{modus tollens}, \ie, denying the consequent~\cite{SPERANZA2010277,sep-logic-ancient}: ``If P then Q. Not Q. Therefore, Not P.''
For example, ``\textit{If something is a intelligent being (P), then it must have the ability to think (Q). Computers cannot think (Not Q). Therefore, computers are not intelligent beings (Not P).}''

To reason about positive propositions, we use \textit{modus ponens} logic, \ie, affirming the antecedent~\cite{sep-logic-ancient}: ``If P then Q. P. Therefore, Q.''
For example, ``\textit{Things with lightweight bodies and strong wing muscles (P) can usually fly (Q). Birds have these physical characteristics (P). Therefore, birds can fly. (Q)}''
Notice that the deduction is not strictly logical but is enough to arrive at commonsense knowledge.

\paragraph{Fact Comparison Prompting}
Deduction emphasizes the intensional aspects of the fact, whereas fact comparison highlights the extensional comparison between counterpart facts~\cite{fitting2006intensional}.
For example, the related fact for ``\textit{lions do not live in the ocean}'' is ``\textit{lions live in the land}''.
A negative fact often comes with a core fact that is true, which has been shown to be useful in explaining why a claim is wrong~\cite{cheng2022unsupervised}.
Therefore, we extend the <input, output> in each example by <input, ``\textit{Related fact}: ...'', output>.
For positive cases, we write a related fact for consistent examples.

\begin{table}[t]
    \centering
    \small
    \begin{tabular}{llcccccc}
    \toprule
        \multirow{2}{*}{\textbf{Model}} & \multirow{2}{*}{\textbf{CoT}}  & \multicolumn{3}{c}{$k=2$ (1:1)} & \multicolumn{3}{c}{$k=10$ (1:1)} \\
        \cmidrule(lr){3-5}
        \cmidrule(lr){6-8}
        & & \pac & \nac & \aac & \pac & \nac & \aac \\
    \midrule
        \multirow{3}{*}{Codex$_\texttt{002}$} & None  & \textbf{96.6} & 38.0 & 67.3 & \textbf{93.2} & 68.8 & 81.0 \\
        & \textit{Deduction} & {86.9} & \textbf{56.6} & {71.7} & {83.5} & {73.0} & {78.3} \\
        & \textit{Fact}  & {92.9} & {53.7} & {\textbf{73.3}} & {86.8} & {\textbf{76.6}} & {\textbf{81.7}} \\
        \midrule
        \multirow{3}{*}{\makecell[c]{Instruct-\\GPT$_\texttt{002}$}} & None & \textbf{92.9} & 51.4 & 72.1 & \textbf{88.9} & 61.4 & 75.1\\
        & \textit{Deduction} & 87.0 & \textbf{57.3} & 72.1 & 84.3 & \textbf{70.7} & \textbf{77.5} \\
        & \textit{Fact}  & 89.1 & 55.5 & \textbf{72.2} & 85.5 & 69.2 & 77.4\\
    \bottomrule
    \end{tabular}
    \caption{Performance on the CG task when enhanced with different types of CoT prompting, \ie, deductive argumentation (\textit{Deduction}) and fact comparison (\textit{Fact}).}
    \label{tab:cot}
\end{table}

\paragraph{Results}

Table~\ref{tab:cot} displays the results of Codex$_\texttt{002}$ and InstructGPT$_\texttt{002}$.
Both CoT instances improve LLMs' performance on \nac, showing the benefit of explicit reasoning for deriving negative knowledge, where different models prefer different rationales.
However, the increase in \nac comes at the expense of a performance drop in \pac.
This is mostly because models previously predicted most of the cases to be positive, making \pac irrationally high.
Overall, these results suggest that, even though LLMs picked up implicit bias during pre-training, it can be overcome by making the reasoning chain explicit.

Nevertheless, deductive reasoning seems to be more rigid about confirming commonsense knowledge with a lower \pac.
This can be attributed to the fact that commonsense knowledge contains exceptions~\cite{allaway2022penguins}, \eg, \textit{birds can fly but penguins can't}.
Thus, LLMs with deductive reasoning may hold concerns about exceptions for confirming a commonsense fact, leading to a significant lower \pac than fact comparison.
We conduct a simple experiment of exceptions in Appendix~\ref{appendix:exception}, which shows that adding adverbs of degree (\eg, \textit{usually}, \textit{generally}) in the texts alleviates the belief conflict, but the problem still exists.

\section{Closing Remarks}
\label{sec:conclusion}

In this study, we explored and quantified the limitations of LLMs in generating texts grounded in negative commonsense knowledge that they seem to know, a phenomenon we term as ``belief conflict''. 
To investigate this, we probe LLMs with a constrained sentence generation (CG) task, coupled with a QA task.
Our experiments demonstrated the existence of the belief conflict in all LLMs when it comes to negative knowledge, which is mostly brought by quantifiable statistical shortcuts such as keywords co-occurrence.
We also see that this can be lessened by giving more in-context examples of negative knowledge or by using a chain-of-thought (CoT) prompting method to explain the explicit reasoning process for deriving negative knowledge.

With the rapid increase of the study on language-based reasoning~\cite{ijcai2020-0537,tafjord-etal-2021-proofwriter,wei2022chain}, there would be cause for concern if LLMs have trouble generating proofs or reasoning steps with negative knowledge.
With all the good scores they achieve at QA tasks, whether they can be trusted with their knowledge expressed during generation, which is one of the most prominent way of human-AI interaction, is still questionable. 
In this sense, the study of negative knowledge creates a good testbed for assessing real language-based reasoning skills for LLMs without the statistical heuristics they memorized.
We hope that the findings in this work could raise the awareness of the community on negative knowledge for LLMs in downstream text generation tasks.


\section*{Limitations}
\label{sec:limitation}
In this work, we highlight that the probing tasks are placed in the commonsense domain that are generally acknowledged by people in most situations.
We do not consider the exceptions of commonsense knowledge, which has gradually drawn some research attentions~\cite{do-pavlick-2021-rotten,allaway2022penguins}.
Exceptions are important for negative knowledge and are widely used in tasks such as argumentation or deductive reasoning.
However, in the experiments, we find that such exceptions might make models generate commonsense statements with uncertain adverbs (\eg, \textit{may}, \textit{some}, etc.) on rare cases.

Another limitation of this work is that the probing task is based only on relational commonsense knowledge from commonsense knowledge bases such as ConceptNet.
We design the keyword-to-sentence task mostly for the purpose of convenient evaluation for text generation, which is notoriously known as difficult.
The probing and evaluation of LLMs' belief about negative knowledge in more complex tasks are beyond the scope of this work, but really interesting and challenging.
Also, other types of knowledge could be studied in a similar way, such as negative social, temporal and spatial knowledge, to name but a few.

In this paper, we identify the belief conflict problem in LLMs through extensive experiments.
Future work could explore more advanced training or prompting-based methods to improve the consistency between a model's belief and its actions (text generation for various tasks), especially for negative knowledge.

\section*{Ethical Statement}

The commonsense knowledge triples from ConceptNet may include offensive and biased sentences, which may also exist in the dataset that we use in this work.
As stated before, the identification of commonsense negative knowledge may slightly vary from people from different cultural and social background when considering exceptions.

\section*{Acknowledgement}
We thank the anonymous reviewers for their valuable comments.
We also thank Siyu Yuan and Jian Xie from Fudan University, and Kexun Zhang, Yujian Liu, Qingxiu Dong and Xuandong Zhao from UC Santa Barbra for their useful suggestions and discussions for the manuscript.
This research is funded by the Science and Technology Commission of Shanghai Municipality Grant (No. 22511105902).

\bibliography{custom}

\begin{thebibliography}{68}
\expandafter\ifx\csname natexlab\endcsname\relax\def\natexlab#1{#1}\fi

\bibitem[{Aggarwal et~al.(2021)Aggarwal, Mandowara, Agrawal, Khandelwal,
  Singla, and Garg}]{aggarwal-etal-2021-explanations}
Shourya Aggarwal, Divyanshu Mandowara, Vishwajeet Agrawal, Dinesh Khandelwal,
  Parag Singla, and Dinesh Garg. 2021.
\newblock \href {https://doi.org/10.18653/v1/2021.acl-long.238} {{E}xplanations
  for {C}ommonsense{QA}: {N}ew {D}ataset and {M}odels}.
\newblock In \emph{Proceedings of the 59th Annual Meeting of the Association
  for Computational Linguistics and the 11th International Joint Conference on
  Natural Language Processing (Volume 1: Long Papers)}, pages 3050--3065,
  Online. Association for Computational Linguistics.

\bibitem[{Allaway et~al.(2022)Allaway, Hwang, Bhagavatula, McKeown, Downey, and
  Choi}]{allaway2022penguins}
Emily Allaway, Jena~D Hwang, Chandra Bhagavatula, Kathleen McKeown, Doug
  Downey, and Yejin Choi. 2022.
\newblock Penguins don't fly: Reasoning about generics through instantiations
  and exceptions.
\newblock \emph{arXiv preprint arXiv:2205.11658}.

\bibitem[{Arnaout et~al.(2021)Arnaout, Razniewski, Weikum, and
  Pan}]{10.1145/3442442.3452339}
Hiba Arnaout, Simon Razniewski, Gerhard Weikum, and Jeff~Z. Pan. 2021.
\newblock \href {https://doi.org/10.1145/3442442.3452339} {Negative knowledge
  for open-world wikidata}.
\newblock In \emph{Companion Proceedings of the Web Conference 2021}, WWW '21,
  page 544–551, New York, NY, USA. Association for Computing Machinery.

\bibitem[{Arnaout et~al.(2022)Arnaout, Razniewski, Weikum, and
  Pan}]{10.1145/3511808.3557484}
Hiba Arnaout, Simon Razniewski, Gerhard Weikum, and Jeff~Z. Pan. 2022.
\newblock \href {https://doi.org/10.1145/3511808.3557484} {Uncommonsense:
  Informative negative knowledge about everyday concepts}.
\newblock In \emph{Proceedings of the 31st ACM International Conference on
  Information \& Knowledge Management}, CIKM '22, page 37–46, New York, NY,
  USA. Association for Computing Machinery.

\bibitem[{Auer et~al.(2007)Auer, Bizer, Kobilarov, Lehmann, Cyganiak, and
  Ives}]{auer2007dbpedia}
S{\"o}ren Auer, Christian Bizer, Georgi Kobilarov, Jens Lehmann, Richard
  Cyganiak, and Zachary Ives. 2007.
\newblock Dbpedia: A nucleus for a web of open data.
\newblock In \emph{The semantic web}, pages 722--735. Springer.

\bibitem[{Barker and Jago(2012)}]{barker2012being}
Stephen Barker and Mark Jago. 2012.
\newblock Being positive about negative facts.
\newblock \emph{Philosophy and Phenomenological research}, pages 117--138.

\bibitem[{Bobzien(2020)}]{sep-logic-ancient}
Susanne Bobzien. 2020.
\newblock {Ancient Logic}.
\newblock In Edward~N. Zalta, editor, \emph{The {Stanford} Encyclopedia of
  Philosophy}, {S}ummer 2020 edition. Metaphysics Research Lab, Stanford
  University.

\bibitem[{Bommasani et~al.(2021)Bommasani, Hudson, Adeli, Altman, Arora, von
  Arx, Bernstein, Bohg, Bosselut, Brunskill
  et~al.}]{bommasani2021opportunities}
Rishi Bommasani, Drew~A Hudson, Ehsan Adeli, Russ Altman, Simran Arora, Sydney
  von Arx, Michael~S Bernstein, Jeannette Bohg, Antoine Bosselut, Emma
  Brunskill, et~al. 2021.
\newblock On the opportunities and risks of foundation models.
\newblock \emph{arXiv preprint arXiv:2108.07258}.

\bibitem[{Branco et~al.(2021)Branco, Branco, Ant{\'o}nio~Rodrigues, and
  Silva}]{branco-etal-2021-shortcutted}
Ruben Branco, Ant{\'o}nio Branco, Jo{\~a}o Ant{\'o}nio~Rodrigues, and
  Jo{\~a}o~Ricardo Silva. 2021.
\newblock \href {https://doi.org/10.18653/v1/2021.emnlp-main.113} {Shortcutted
  commonsense: Data spuriousness in deep learning of commonsense reasoning}.
\newblock In \emph{Proceedings of the 2021 Conference on Empirical Methods in
  Natural Language Processing}, pages 1504--1521, Online and Punta Cana,
  Dominican Republic. Association for Computational Linguistics.

\bibitem[{Brown et~al.(2020)Brown, Mann, Ryder, Subbiah, Kaplan, Dhariwal,
  Neelakantan, Shyam, Sastry, Askell, Agarwal, Herbert-Voss, Krueger, Henighan,
  Child, Ramesh, Ziegler, Wu, Winter, Hesse, Chen, Sigler, Litwin, Gray, Chess,
  Clark, Berner, McCandlish, Radford, Sutskever, and
  Amodei}]{NEURIPS2020_1457c0d6}
Tom Brown, Benjamin Mann, Nick Ryder, Melanie Subbiah, Jared~D Kaplan, Prafulla
  Dhariwal, Arvind Neelakantan, Pranav Shyam, Girish Sastry, Amanda Askell,
  Sandhini Agarwal, Ariel Herbert-Voss, Gretchen Krueger, Tom Henighan, Rewon
  Child, Aditya Ramesh, Daniel Ziegler, Jeffrey Wu, Clemens Winter, Chris
  Hesse, Mark Chen, Eric Sigler, Mateusz Litwin, Scott Gray, Benjamin Chess,
  Jack Clark, Christopher Berner, Sam McCandlish, Alec Radford, Ilya Sutskever,
  and Dario Amodei. 2020.
\newblock \href
  {https://proceedings.neurips.cc/paper/2020/file/1457c0d6bfcb4967418bfb8ac142f64a-Paper.pdf}
  {Language models are few-shot learners}.
\newblock In \emph{Advances in Neural Information Processing Systems},
  volume~33, pages 1877--1901. Curran Associates, Inc.

\bibitem[{Camburu et~al.(2018)Camburu, Rockt\"{a}schel, Lukasiewicz, and
  Blunsom}]{NEURIPS2018_4c7a167b}
Oana-Maria Camburu, Tim Rockt\"{a}schel, Thomas Lukasiewicz, and Phil Blunsom.
  2018.
\newblock \href
  {https://proceedings.neurips.cc/paper/2018/file/4c7a167bb329bd92580a99ce422d6fa6-Paper.pdf}
  {e-snli: Natural language inference with natural language explanations}.
\newblock In \emph{Advances in Neural Information Processing Systems},
  volume~31. Curran Associates, Inc.

\bibitem[{Cao et~al.(2021)Cao, Lin, Han, Sun, Yan, Liao, Xue, and
  Xu}]{cao-etal-2021-knowledgeable}
Boxi Cao, Hongyu Lin, Xianpei Han, Le~Sun, Lingyong Yan, Meng Liao, Tong Xue,
  and Jin Xu. 2021.
\newblock \href {https://doi.org/10.18653/v1/2021.acl-long.146} {Knowledgeable
  or educated guess? revisiting language models as knowledge bases}.
\newblock In \emph{Proceedings of the 59th Annual Meeting of the Association
  for Computational Linguistics and the 11th International Joint Conference on
  Natural Language Processing (Volume 1: Long Papers)}, pages 1860--1874,
  Online. Association for Computational Linguistics.

\bibitem[{Chen et~al.(2022)Chen, Xu, Fu, Shi, Li, Zhang, Sun, Li, Xiao, and
  Zhou}]{chen-etal-2022-e}
Jiangjie Chen, Rui Xu, Ziquan Fu, Wei Shi, Zhongqiao Li, Xinbo Zhang, Changzhi
  Sun, Lei Li, Yanghua Xiao, and Hao Zhou. 2022.
\newblock \href {https://doi.org/10.18653/v1/2022.findings-acl.311} {{E}-{KAR}:
  A benchmark for rationalizing natural language analogical reasoning}.
\newblock In \emph{Findings of the Association for Computational Linguistics:
  ACL 2022}, pages 3941--3955, Dublin, Ireland. Association for Computational
  Linguistics.

\bibitem[{Chen et~al.(2021)Chen, Tworek, Jun, Yuan, Pinto, Kaplan, Edwards,
  Burda, Joseph, Brockman et~al.}]{chen2021evaluating}
Mark Chen, Jerry Tworek, Heewoo Jun, Qiming Yuan, Henrique Ponde de~Oliveira
  Pinto, Jared Kaplan, Harri Edwards, Yuri Burda, Nicholas Joseph, Greg
  Brockman, et~al. 2021.
\newblock Evaluating large language models trained on code.
\newblock \emph{arXiv preprint arXiv:2107.03374}.

\bibitem[{Cheng et~al.(2022)Cheng, Wu, Chen, Li, Liu, and
  Kong}]{cheng2022unsupervised}
Sijie Cheng, Zhiyong Wu, Jiangjie Chen, Zhixing Li, Yang Liu, and Lingpeng
  Kong. 2022.
\newblock Unsupervised explanation generation via correct instantiations.
\newblock \emph{arXiv preprint arXiv:2211.11160}.

\bibitem[{Chowdhery et~al.(2022)Chowdhery, Narang, Devlin, Bosma, Mishra,
  Roberts, Barham, Chung, Sutton, Gehrmann et~al.}]{chowdhery2022palm}
Aakanksha Chowdhery, Sharan Narang, Jacob Devlin, Maarten Bosma, Gaurav Mishra,
  Adam Roberts, Paul Barham, Hyung~Won Chung, Charles Sutton, Sebastian
  Gehrmann, et~al. 2022.
\newblock Palm: Scaling language modeling with pathways.
\newblock \emph{arXiv preprint arXiv:2204.02311}.

\bibitem[{Chung et~al.(2022)Chung, Hou, Longpre, Zoph, Tay, Fedus, Li, Wang,
  Dehghani, Brahma et~al.}]{chung2022scaling}
Hyung~Won Chung, Le~Hou, Shayne Longpre, Barret Zoph, Yi~Tay, William Fedus,
  Eric Li, Xuezhi Wang, Mostafa Dehghani, Siddhartha Brahma, et~al. 2022.
\newblock Scaling instruction-finetuned language models.
\newblock \emph{arXiv preprint arXiv:2210.11416}.

\bibitem[{Clark et~al.(2020)Clark, Tafjord, and Richardson}]{ijcai2020-0537}
Peter Clark, Oyvind Tafjord, and Kyle Richardson. 2020.
\newblock \href {https://doi.org/10.24963/ijcai.2020/537} {Transformers as soft
  reasoners over language}.
\newblock In \emph{Proceedings of the Twenty-Ninth International Joint
  Conference on Artificial Intelligence, {IJCAI-20}}, pages 3882--3890.
  International Joint Conferences on Artificial Intelligence Organization.
\newblock Main track.

\bibitem[{Devlin et~al.(2019)Devlin, Chang, Lee, and
  Toutanova}]{devlin-etal-2019-bert}
Jacob Devlin, Ming-Wei Chang, Kenton Lee, and Kristina Toutanova. 2019.
\newblock \href {https://doi.org/10.18653/v1/N19-1423} {{BERT}: Pre-training of
  deep bidirectional transformers for language understanding}.
\newblock In \emph{Proceedings of the 2019 Conference of the North {A}merican
  Chapter of the Association for Computational Linguistics: Human Language
  Technologies, Volume 1 (Long and Short Papers)}, pages 4171--4186,
  Minneapolis, Minnesota. Association for Computational Linguistics.

\bibitem[{Do and Pavlick(2021)}]{do-pavlick-2021-rotten}
Nam Do and Ellie Pavlick. 2021.
\newblock \href {https://doi.org/10.18653/v1/2021.findings-acl.181} {Are rotten
  apples edible? challenging commonsense inference ability with exceptions}.
\newblock In \emph{Findings of the Association for Computational Linguistics:
  ACL-IJCNLP 2021}, pages 2061--2073, Online. Association for Computational
  Linguistics.

\bibitem[{Ettinger(2020)}]{ettinger-2020-bert}
Allyson Ettinger. 2020.
\newblock \href {https://doi.org/10.1162/tacl_a_00298} {What {BERT} is not:
  Lessons from a new suite of psycholinguistic diagnostics for language
  models}.
\newblock \emph{Transactions of the Association for Computational Linguistics},
  8:34--48.

\bibitem[{Fitting(2006)}]{fitting2006intensional}
Melvin Fitting. 2006.
\newblock Intensional logic.

\bibitem[{Gao et~al.(2021)Gao, Yao, and Chen}]{gao-etal-2021-simcse}
Tianyu Gao, Xingcheng Yao, and Danqi Chen. 2021.
\newblock \href {https://doi.org/10.18653/v1/2021.emnlp-main.552} {{S}im{CSE}:
  Simple contrastive learning of sentence embeddings}.
\newblock In \emph{Proceedings of the 2021 Conference on Empirical Methods in
  Natural Language Processing}, pages 6894--6910, Online and Punta Cana,
  Dominican Republic. Association for Computational Linguistics.

\bibitem[{Gubelmann and Handschuh(2022)}]{gubelmann-handschuh-2022-context}
Reto Gubelmann and Siegfried Handschuh. 2022.
\newblock \href {https://doi.org/10.18653/v1/2022.acl-long.315} {Context
  matters: A pragmatic study of {PLM}s{'} negation understanding}.
\newblock In \emph{Proceedings of the 60th Annual Meeting of the Association
  for Computational Linguistics (Volume 1: Long Papers)}, pages 4602--4621,
  Dublin, Ireland. Association for Computational Linguistics.

\bibitem[{Gururangan et~al.(2018)Gururangan, Swayamdipta, Levy, Schwartz,
  Bowman, and Smith}]{gururangan-etal-2018-annotation}
Suchin Gururangan, Swabha Swayamdipta, Omer Levy, Roy Schwartz, Samuel Bowman,
  and Noah~A. Smith. 2018.
\newblock \href {https://doi.org/10.18653/v1/N18-2017} {Annotation artifacts in
  natural language inference data}.
\newblock In \emph{Proceedings of the 2018 Conference of the North {A}merican
  Chapter of the Association for Computational Linguistics: Human Language
  Technologies, Volume 2 (Short Papers)}, pages 107--112, New Orleans,
  Louisiana. Association for Computational Linguistics.

\bibitem[{Horn and Wansing(2022)}]{sep-negation}
Laurence~R. Horn and Heinrich Wansing. 2022.
\newblock {Negation}.
\newblock In Edward~N. Zalta and Uri Nodelman, editors, \emph{The {Stanford}
  Encyclopedia of Philosophy}, {W}inter 2022 edition. Metaphysics Research Lab,
  Stanford University.

\bibitem[{Hossain et~al.(2022)Hossain, Chinnappa, and
  Blanco}]{hossain-etal-2022-analysis}
Md~Mosharaf Hossain, Dhivya Chinnappa, and Eduardo Blanco. 2022.
\newblock \href {https://doi.org/10.18653/v1/2022.acl-short.81} {An analysis of
  negation in natural language understanding corpora}.
\newblock In \emph{Proceedings of the 60th Annual Meeting of the Association
  for Computational Linguistics (Volume 2: Short Papers)}, pages 716--723,
  Dublin, Ireland. Association for Computational Linguistics.

\bibitem[{Hosseini et~al.(2021)Hosseini, Reddy, Bahdanau, Hjelm, Sordoni, and
  Courville}]{hosseini-etal-2021-understanding}
Arian Hosseini, Siva Reddy, Dzmitry Bahdanau, R~Devon Hjelm, Alessandro
  Sordoni, and Aaron Courville. 2021.
\newblock \href {https://doi.org/10.18653/v1/2021.naacl-main.102}
  {Understanding by understanding not: Modeling negation in language models}.
\newblock In \emph{Proceedings of the 2021 Conference of the North American
  Chapter of the Association for Computational Linguistics: Human Language
  Technologies}, pages 1301--1312, Online. Association for Computational
  Linguistics.

\bibitem[{Jang et~al.(2022)Jang, Ye, and Seo}]{jang2022can}
Joel Jang, Seonghyeon Ye, and Minjoon Seo. 2022.
\newblock Can large language models truly understand prompts? a case study with
  negated prompts.
\newblock \emph{arXiv preprint arXiv:2209.12711}.

\bibitem[{Jung et~al.(2022)Jung, Qin, Welleck, Brahman, Bhagavatula, Bras, and
  Choi}]{jung2022maieutic}
Jaehun Jung, Lianhui Qin, Sean Welleck, Faeze Brahman, Chandra Bhagavatula,
  Ronan~Le Bras, and Yejin Choi. 2022.
\newblock Maieutic prompting: Logically consistent reasoning with recursive
  explanations.
\newblock \emph{arXiv preprint arXiv:2205.11822}.

\bibitem[{Kassner and Sch{\"u}tze(2020)}]{kassner-schutze-2020-negated}
Nora Kassner and Hinrich Sch{\"u}tze. 2020.
\newblock \href {https://doi.org/10.18653/v1/2020.acl-main.698} {Negated and
  misprimed probes for pretrained language models: Birds can talk, but cannot
  fly}.
\newblock In \emph{Proceedings of the 58th Annual Meeting of the Association
  for Computational Linguistics}, pages 7811--7818, Online. Association for
  Computational Linguistics.

\bibitem[{Kassner et~al.(2021)Kassner, Tafjord, Sch{\"u}tze, and
  Clark}]{kassner-etal-2021-beliefbank}
Nora Kassner, Oyvind Tafjord, Hinrich Sch{\"u}tze, and Peter Clark. 2021.
\newblock \href {https://doi.org/10.18653/v1/2021.emnlp-main.697}
  {{B}elief{B}ank: Adding memory to a pre-trained language model for a
  systematic notion of belief}.
\newblock In \emph{Proceedings of the 2021 Conference on Empirical Methods in
  Natural Language Processing}, pages 8849--8861, Online and Punta Cana,
  Dominican Republic. Association for Computational Linguistics.

\bibitem[{Lai et~al.(2021)Lai, Zhang, Feng, Huang, and
  Zhao}]{lai-etal-2021-machine}
Yuxuan Lai, Chen Zhang, Yansong Feng, Quzhe Huang, and Dongyan Zhao. 2021.
\newblock \href {https://doi.org/10.18653/v1/2021.findings-acl.85} {Why machine
  reading comprehension models learn shortcuts?}
\newblock In \emph{Findings of the Association for Computational Linguistics:
  ACL-IJCNLP 2021}, pages 989--1002, Online. Association for Computational
  Linguistics.

\bibitem[{Lin et~al.(2020)Lin, Zhou, Shen, Zhou, Bhagavatula, Choi, and
  Ren}]{lin-etal-2020-commongen}
Bill~Yuchen Lin, Wangchunshu Zhou, Ming Shen, Pei Zhou, Chandra Bhagavatula,
  Yejin Choi, and Xiang Ren. 2020.
\newblock \href {https://doi.org/10.18653/v1/2020.findings-emnlp.165}
  {{C}ommon{G}en: A constrained text generation challenge for generative
  commonsense reasoning}.
\newblock In \emph{Findings of the Association for Computational Linguistics:
  EMNLP 2020}, pages 1823--1840, Online. Association for Computational
  Linguistics.

\bibitem[{Liu et~al.(2021)Liu, Wan, He, Peng, and Yu}]{Liu_Wan_He_Peng_Yu_2021}
Ye~Liu, Yao Wan, Lifang He, Hao Peng, and Philip~S. Yu. 2021.
\newblock \href {https://doi.org/10.1609/aaai.v35i7.16796} {Kg-bart: Knowledge
  graph-augmented bart for generative commonsense reasoning}.
\newblock \emph{Proceedings of the AAAI Conference on Artificial Intelligence},
  35(7):6418--6425.

\bibitem[{Liu et~al.(2019)Liu, Ott, Goyal, Du, Joshi, Chen, Levy, Lewis,
  Zettlemoyer, and Stoyanov}]{liu2019roberta}
Yinhan Liu, Myle Ott, Naman Goyal, Jingfei Du, Mandar Joshi, Danqi Chen, Omer
  Levy, Mike Lewis, Luke Zettlemoyer, and Veselin Stoyanov. 2019.
\newblock Roberta: A robustly optimized bert pretraining approach.
\newblock \emph{arXiv preprint arXiv:1907.11692}.

\bibitem[{MacDonald(1965)}]{1020071ar}
Charles MacDonald. 1965.
\newblock \href {https://doi.org/https://doi.org/10.7202/1020071ar} {The role
  of negation in human knowledge}.
\newblock \emph{Laval théologique et philosophique}, 21(1):80--114.

\bibitem[{Min et~al.(2022{\natexlab{a}})Min, Lewis, Zettlemoyer, and
  Hajishirzi}]{min-etal-2022-metaicl}
Sewon Min, Mike Lewis, Luke Zettlemoyer, and Hannaneh Hajishirzi.
  2022{\natexlab{a}}.
\newblock \href {https://doi.org/10.18653/v1/2022.naacl-main.201} {{M}eta{ICL}:
  Learning to learn in context}.
\newblock In \emph{Proceedings of the 2022 Conference of the North American
  Chapter of the Association for Computational Linguistics: Human Language
  Technologies}, pages 2791--2809, Seattle, United States. Association for
  Computational Linguistics.

\bibitem[{Min et~al.(2022{\natexlab{b}})Min, Lyu, Holtzman, Artetxe, Lewis,
  Hajishirzi, and Zettlemoyer}]{min2022rethinking}
Sewon Min, Xinxi Lyu, Ari Holtzman, Mikel Artetxe, Mike Lewis, Hannaneh
  Hajishirzi, and Luke Zettlemoyer. 2022{\natexlab{b}}.
\newblock Rethinking the role of demonstrations: What makes in-context learning
  work?
\newblock \emph{arXiv preprint arXiv:2202.12837}.

\bibitem[{Minsky(1997)}]{Minsky1997NegativeE}
Marvin Minsky. 1997.
\newblock Negative expertise.

\bibitem[{Molnar(2000)}]{molnar2000truthmakers}
George Molnar. 2000.
\newblock Truthmakers for negative truths.
\newblock \emph{Australasian Journal of philosophy}, 78(1):72--86.

\bibitem[{OpenAI(2022)}]{openai2022chatgpt}
OpenAI. 2022.
\newblock \href {https://openai.com/blog/chatgpt} {Chatgpt}.

\bibitem[{Ouyang et~al.(2022)Ouyang, Wu, Jiang, Almeida, Wainwright, Mishkin,
  Zhang, Agarwal, Slama, Gray, Schulman, Hilton, Kelton, Miller, Simens,
  Askell, Welinder, Christiano, Leike, and Lowe}]{ouyang2022training}
Long Ouyang, Jeffrey Wu, Xu~Jiang, Diogo Almeida, Carroll Wainwright, Pamela
  Mishkin, Chong Zhang, Sandhini Agarwal, Katarina Slama, Alex Gray, John
  Schulman, Jacob Hilton, Fraser Kelton, Luke Miller, Maddie Simens, Amanda
  Askell, Peter Welinder, Paul Christiano, Jan Leike, and Ryan Lowe. 2022.
\newblock \href {https://openreview.net/forum?id=TG8KACxEON} {Training language
  models to follow instructions with human feedback}.
\newblock In \emph{Advances in Neural Information Processing Systems}.

\bibitem[{Petroni et~al.(2020)Petroni, Lewis, Piktus, Rockt{\"a}schel, Wu,
  Miller, and Riedel}]{petroni2020how}
Fabio Petroni, Patrick Lewis, Aleksandra Piktus, Tim Rockt{\"a}schel, Yuxiang
  Wu, Alexander~H. Miller, and Sebastian Riedel. 2020.
\newblock \href {https://doi.org/10.24432/C5201W} {How context affects language
  models' factual predictions}.
\newblock In \emph{Automated Knowledge Base Construction}.

\bibitem[{Petroni et~al.(2019)Petroni, Rockt{\"a}schel, Riedel, Lewis, Bakhtin,
  Wu, and Miller}]{petroni-etal-2019-language}
Fabio Petroni, Tim Rockt{\"a}schel, Sebastian Riedel, Patrick Lewis, Anton
  Bakhtin, Yuxiang Wu, and Alexander Miller. 2019.
\newblock \href {https://doi.org/10.18653/v1/D19-1250} {Language models as
  knowledge bases?}
\newblock In \emph{Proceedings of the 2019 Conference on Empirical Methods in
  Natural Language Processing and the 9th International Joint Conference on
  Natural Language Processing (EMNLP-IJCNLP)}, pages 2463--2473, Hong Kong,
  China. Association for Computational Linguistics.

\bibitem[{Radford et~al.(2019)Radford, Wu, Child, Luan, Amodei, and
  Sutskever}]{radford2019language}
Alec Radford, Jeffrey Wu, Rewon Child, David Luan, Dario Amodei, and Ilya
  Sutskever. 2019.
\newblock Language models are unsupervised multitask learners.
\newblock \emph{OpenAI Blog}, 1(8):9.

\bibitem[{Raffel et~al.(2020)Raffel, Shazeer, Roberts, Lee, Narang, Matena,
  Zhou, Li, and Liu}]{raffel2020exploring}
Colin Raffel, Noam Shazeer, Adam Roberts, Katherine Lee, Sharan Narang, Michael
  Matena, Yanqi Zhou, Wei Li, and Peter~J Liu. 2020.
\newblock Exploring the limits of transfer learning with a unified text-to-text
  transformer.
\newblock \emph{Journal of Machine Learning Research}, 21(140):1--67.

\bibitem[{Rajpurkar et~al.(2016)Rajpurkar, Zhang, Lopyrev, and
  Liang}]{rajpurkar-etal-2016-squad}
Pranav Rajpurkar, Jian Zhang, Konstantin Lopyrev, and Percy Liang. 2016.
\newblock \href {https://doi.org/10.18653/v1/D16-1264} {{SQ}u{AD}: 100,000+
  questions for machine comprehension of text}.
\newblock In \emph{Proceedings of the 2016 Conference on Empirical Methods in
  Natural Language Processing}, pages 2383--2392, Austin, Texas. Association
  for Computational Linguistics.

\bibitem[{Ravichander et~al.(2022)Ravichander, Gardner, and
  Marasovi{\'c}}]{ravichander2022condaqa}
Abhilasha Ravichander, Matt Gardner, and Ana Marasovi{\'c}. 2022.
\newblock Condaqa: A contrastive reading comprehension dataset for reasoning
  about negation.
\newblock \emph{arXiv preprint arXiv:2211.00295}.

\bibitem[{Reiter(2019)}]{reiter2019natural}
Ehud Reiter. 2019.
\newblock Natural language generation challenges for explainable ai.
\newblock \emph{arXiv preprint arXiv:1911.08794}.

\bibitem[{Richardson et~al.(2022)Richardson, Tamari, Sultan, Tsarfaty, Shahaf,
  and Sabharwal}]{richardson2022breakpoint}
Kyle Richardson, Ronen Tamari, Oren Sultan, Reut Tsarfaty, Dafna Shahaf, and
  Ashish Sabharwal. 2022.
\newblock Breakpoint transformers for modeling and tracking intermediate
  beliefs.
\newblock \emph{arXiv preprint arXiv:2211.07950}.

\bibitem[{Rubin et~al.(2022)Rubin, Herzig, and
  Berant}]{rubin-etal-2022-learning}
Ohad Rubin, Jonathan Herzig, and Jonathan Berant. 2022.
\newblock \href {https://doi.org/10.18653/v1/2022.naacl-main.191} {Learning to
  retrieve prompts for in-context learning}.
\newblock In \emph{Proceedings of the 2022 Conference of the North American
  Chapter of the Association for Computational Linguistics: Human Language
  Technologies}, pages 2655--2671, Seattle, United States. Association for
  Computational Linguistics.

\bibitem[{Safavi et~al.(2021)Safavi, Zhu, and
  Koutra}]{safavi-etal-2021-negater}
Tara Safavi, Jing Zhu, and Danai Koutra. 2021.
\newblock \href {https://doi.org/10.18653/v1/2021.emnlp-main.456} {{N}egat{ER}:
  {U}nsupervised {D}iscovery of {N}egatives in {C}ommonsense {K}nowledge
  {B}ases}.
\newblock In \emph{Proceedings of the 2021 Conference on Empirical Methods in
  Natural Language Processing}, pages 5633--5646, Online and Punta Cana,
  Dominican Republic. Association for Computational Linguistics.

\bibitem[{Scao et~al.(2022)Scao, Fan, Akiki, Pavlick, Ili{\'c}, Hesslow,
  Castagn{\'e}, Luccioni, Yvon, Gall{\'e} et~al.}]{scao2022bloom}
Teven~Le Scao, Angela Fan, Christopher Akiki, Ellie Pavlick, Suzana Ili{\'c},
  Daniel Hesslow, Roman Castagn{\'e}, Alexandra~Sasha Luccioni, Fran{\c{c}}ois
  Yvon, Matthias Gall{\'e}, et~al. 2022.
\newblock Bloom: A 176b-parameter open-access multilingual language model.
\newblock \emph{arXiv preprint arXiv:2211.05100}.

\bibitem[{Singh et~al.(2002)Singh, Lin, Mueller, Lim, Perkins, and
  Li~Zhu}]{10.1007/3-540-36124-3_77}
Push Singh, Thomas Lin, Erik~T. Mueller, Grace Lim, Travell Perkins, and Wan
  Li~Zhu. 2002.
\newblock Open mind common sense: Knowledge acquisition from the general
  public.
\newblock In \emph{On the Move to Meaningful Internet Systems 2002: CoopIS,
  DOA, and ODBASE}, pages 1223--1237, Berlin, Heidelberg. Springer Berlin
  Heidelberg.

\bibitem[{Speer et~al.(2017)Speer, Chin, and Havasi}]{Speer_Chin_Havasi_2017}
Robyn Speer, Joshua Chin, and Catherine Havasi. 2017.
\newblock \href {https://doi.org/10.1609/aaai.v31i1.11164} {Conceptnet 5.5: An
  open multilingual graph of general knowledge}.
\newblock \emph{Proceedings of the AAAI Conference on Artificial Intelligence},
  31(1).

\bibitem[{Speranza and Horn(2010)}]{SPERANZA2010277}
J.L. Speranza and Laurence~R. Horn. 2010.
\newblock \href {https://doi.org/https://doi.org/10.1016/j.jal.2010.04.001} {A
  brief history of negation}.
\newblock \emph{Journal of Applied Logic}, 8(3):277--301.

\bibitem[{Sumers et~al.(2021)Sumers, Hawkins, Ho, and
  Griffiths}]{sumers2021extending}
Theodore~R Sumers, Robert~D Hawkins, Mark~K Ho, and Thomas~L Griffiths. 2021.
\newblock Extending rational models of communication from beliefs to actions.
\newblock \emph{arXiv preprint arXiv:2105.11950}.

\bibitem[{Tafjord et~al.(2021)Tafjord, Dalvi, and
  Clark}]{tafjord-etal-2021-proofwriter}
Oyvind Tafjord, Bhavana Dalvi, and Peter Clark. 2021.
\newblock \href {https://doi.org/10.18653/v1/2021.findings-acl.317}
  {{P}roof{W}riter: Generating implications, proofs, and abductive statements
  over natural language}.
\newblock In \emph{Findings of the Association for Computational Linguistics:
  ACL-IJCNLP 2021}, pages 3621--3634, Online. Association for Computational
  Linguistics.

\bibitem[{Tafjord et~al.(2022)Tafjord, Mishra, and Clark}]{tafjord2022entailer}
Oyvind Tafjord, Bhavana~Dalvi Mishra, and Peter Clark. 2022.
\newblock Entailer: Answering questions with faithful and truthful chains of
  reasoning.
\newblock \emph{arXiv preprint arXiv:2210.12217}.

\bibitem[{Talmor et~al.(2019)Talmor, Herzig, Lourie, and
  Berant}]{talmor-etal-2019-commonsenseqa}
Alon Talmor, Jonathan Herzig, Nicholas Lourie, and Jonathan Berant. 2019.
\newblock \href {https://doi.org/10.18653/v1/N19-1421} {{C}ommonsense{QA}: A
  question answering challenge targeting commonsense knowledge}.
\newblock In \emph{Proceedings of the 2019 Conference of the North {A}merican
  Chapter of the Association for Computational Linguistics: Human Language
  Technologies, Volume 1 (Long and Short Papers)}, pages 4149--4158,
  Minneapolis, Minnesota. Association for Computational Linguistics.

\bibitem[{Tian et~al.(2022)Tian, Cao, Zhang, and Xing}]{tian2022debiasing}
Bing Tian, Yixin Cao, Yong Zhang, and Chunxiao Xing. 2022.
\newblock Debiasing nlu models via causal intervention and counterfactual
  reasoning.

\bibitem[{Vrande{\v{c}}i{\'c} and Kr{\"o}tzsch(2014)}]{vrandevcic2014wikidata}
Denny Vrande{\v{c}}i{\'c} and Markus Kr{\"o}tzsch. 2014.
\newblock Wikidata: a free collaborative knowledgebase.
\newblock \emph{Communications of the ACM}, 57(10):78--85.

\bibitem[{Wang et~al.(2022)Wang, Wei, Schuurmans, Le, Chi, and
  Zhou}]{wang2022self}
Xuezhi Wang, Jason Wei, Dale Schuurmans, Quoc Le, Ed~Chi, and Denny Zhou. 2022.
\newblock Self-consistency improves chain of thought reasoning in language
  models.
\newblock \emph{arXiv preprint arXiv:2203.11171}.

\bibitem[{Wei et~al.(2022{\natexlab{a}})Wei, Tay, Bommasani, Raffel, Zoph,
  Borgeaud, Yogatama, Bosma, Zhou, Metzler, Chi, Hashimoto, Vinyals, Liang,
  Dean, and Fedus}]{wei2022emergent}
Jason Wei, Yi~Tay, Rishi Bommasani, Colin Raffel, Barret Zoph, Sebastian
  Borgeaud, Dani Yogatama, Maarten Bosma, Denny Zhou, Donald Metzler, Ed~H.
  Chi, Tatsunori Hashimoto, Oriol Vinyals, Percy Liang, Jeff Dean, and William
  Fedus. 2022{\natexlab{a}}.
\newblock \href {https://openreview.net/forum?id=yzkSU5zdwD} {Emergent
  abilities of large language models}.
\newblock \emph{Transactions on Machine Learning Research}.
\newblock Survey Certification.

\bibitem[{Wei et~al.(2022{\natexlab{b}})Wei, Wang, Schuurmans, Bosma, brian
  ichter, Xia, Chi, Le, and Zhou}]{wei2022chain}
Jason Wei, Xuezhi Wang, Dale Schuurmans, Maarten Bosma, brian ichter, Fei Xia,
  Ed~H. Chi, Quoc~V Le, and Denny Zhou. 2022{\natexlab{b}}.
\newblock \href {https://openreview.net/forum?id=_VjQlMeSB_J} {Chain of thought
  prompting elicits reasoning in large language models}.
\newblock In \emph{Advances in Neural Information Processing Systems}.

\bibitem[{Welleck et~al.(2020)Welleck, Kulikov, Roller, Dinan, Cho, and
  Weston}]{Welleck2020Neural}
Sean Welleck, Ilia Kulikov, Stephen Roller, Emily Dinan, Kyunghyun Cho, and
  Jason Weston. 2020.
\newblock \href {https://openreview.net/forum?id=SJeYe0NtvH} {Neural text
  generation with unlikelihood training}.
\newblock In \emph{International Conference on Learning Representations}.

\bibitem[{Yu et~al.(2022)Yu, Zhu, Li, Hu, Wang, Ji, and Jiang}]{yu2022survey}
Wenhao Yu, Chenguang Zhu, Zaitang Li, Zhiting Hu, Qingyun Wang, Heng Ji, and
  Meng Jiang. 2022.
\newblock A survey of knowledge-enhanced text generation.
\newblock \emph{ACM Computing Surveys (CSUR)}.

\end{thebibliography}
\bibliographystyle{acl_natbib}

\clearpage
\begin{appendix}
\label{sec:appendix}

\section{Demonstrations for In-Context Learning}

\subsection{Manually-written Examples for In-Context Learning}
\label{appendix:demo4probing}

Some of the manually designed examples are shown in Table~\ref{tab:man_examples}.

\subsection{Example Prompts for the Probing Tasks}
\label{appendix:icl_ex}

\begin{table}[t]
    \small
    \centering
    \begin{tabular}{c}
    \toprule
        \rowcolor[gray]{0.95} 
    \textbf{Task 1: Boolean Question Answering (QA)} \\
    \makecell[l{p{7.5cm}}]{\textit{Answer the commonsense questions with yes or no:} \\
    \textcolor{gray}{/* Examples */} \\
    \textbf{Question}: can birds fly? \\
    \textbf{Answer}: yes \\
    \#\#\# \\
    \textbf{Question}: is water spicy? \\
    \textbf{Answer}: no \\
    \textcolor{gray}{/* Test data */} \\
    \textbf{Question}: are needles used for writing? \\
    \textbf{Answer}: \underline{no}} \\
    
    \midrule

    \rowcolor[gray]{0.95} 
    \textbf{Task 2: Constrained Sentence Generation (CG)} \\
    
    \makecell[l{p{7.5cm}}]{\textit{Write a short and factual sentence according to commonsense based on the keywords:} \\
    \textcolor{gray}{/* Examples */} \\
    \textbf{Keywords}: birds, capable of, fly \\
    \textbf{Sentence}: birds can fly. \\
    \#\#\# \\
    \textbf{Keywords}: water, has property, spicy \\
    \textbf{Sentence}: water isn't spicy. \\
    \textcolor{gray}{/* Test data */} \\
    \textbf{Keywords}: needles, used for, writing \\
    \textbf{Sentence}: \underline{needles are not used for writing.}} \\
    \bottomrule
    \end{tabular}
    \caption{Example prompts of the two probing tasks for in-context learning, which consists of a task instruction at the beginning and several in-context examples.
    \underline{Underlined texts} denote the model completion.
    }
    \label{tab:icl_ex}
\end{table}

The task inputs to the LLMs are presented in Table~\ref{tab:icl_ex}.
Note that \textit{instructions} can be replaced by others.
LLMs with in-context learning are known to be sensitive to the wording and examples in the prompts~\cite{min2022rethinking}.
Therefore, we manually write 4 interchangeable instructions for each probing tasks.
For the QA task, the instructions include:
\begin{enumerate}[noitemsep]
    \item \textit{Answer the commonsense questions with yes or no.}
    \item \textit{Choose ``yes'' or ``no'' to indicate whether you agree or disagree with the commonsense questions.}
    \item \textit{Respond to the questions using ``yes'' or ``no''.}
    \item \textit{Indicate whether the commonsense questions are correct or incorrect by writing ``yes'' or ``no''.}
\end{enumerate}

For the CG task, the instructions include:
\begin{enumerate}[noitemsep]
    \item \textit{Write a short and factual sentence according to commonsense based on the keywords:}
    \item \textit{Use the keywords to create a short and factual sentence that accurately reflects commonsense knowledge.}
    \item \textit{Create a short, factual sentence based on the keywords and what is generally accepted as true.}
    \item \textit{Construct a factual and concise statement based on the provided keywords and commonsense knowledge.}
\end{enumerate}

\section{Additional Results}

\subsection{Sensitivity to Temperature Tuning}
\label{appendix:temperature}

\begin{figure}[t]
    \centering
	\subfigure[Results (\%) on QA.] {
    \label{fig:temperature_qa} 
		\pgfplotsset{width=0.345\linewidth,height=0.33\linewidth,compat=1.5,scale only axis}
\footnotesize
\begin{tikzpicture}
\begin{axis}[
    xlabel={temperature},
    xmin=-0.1, xmax=1.1,
    ymin=0, ymax=100,
    xtick={0, 0.2, 0.4, 0.6, 0.8, 1},
    legend pos=south east,
    ymajorgrids=true,
    xmajorgrids=true,
    grid style=dashed,
    x label style={at={(axis description cs:0.5,-0.125)},anchor=north},
    legend style={nodes={scale=0.7, transform shape},font=\footnotesize}
]
\addplot[
    color=NavyBlue,
    mark=o,
    mark size=1.5pt,
    ]
    coordinates {
    (0, 84.1)
    (0.2, 84.75)
    (0.4, 84.7)
    (0.6, 84.7)
    (0.8, 84.65)
    (1, 84.75)
    };
    \addlegendentry{\pac}
\addplot[
    color=Maroon,
    mark=triangle,
    mark size=2pt,
    ]
    coordinates {
    (0, 84.7)
    (0.2, 84.4)
    (0.4, 84.3)
    (0.6, 84.3)
    (0.8, 84.3)
    (1, 84.35)
    };
    \addlegendentry{\nac}
\addplot[
    color=Green,
    mark=diamond,
    mark size=2pt,
    ]
    coordinates {
    (0, 84.4)
    (0.2, 84.575)
    (0.4, 84.5)
    (0.6, 84.5)
    (0.8, 84.475)
    (1, 84.55)
    };
    \addlegendentry{\aac}
\end{axis}
\end{tikzpicture}

}
	\subfigure[Results (\%) on CG.] { 
	\label{fig:temperature_cg} 
		\pgfplotsset{width=0.345\linewidth,height=0.33\linewidth,compat=1.5,scale only axis}
\footnotesize
\begin{tikzpicture}
\begin{axis}[
    xlabel={temperature},
    xmin=-0.1, xmax=1.1,
    ymin=0, ymax=100,
    xtick={0, 0.2, 0.4, 0.6, 0.8, 1},
    legend pos=south east,
    ymajorgrids=true,
    xmajorgrids=true,
    grid style=dashed,
    x label style={at={(axis description cs:0.5,-0.125)},anchor=north},
    legend style={nodes={scale=0.7, transform shape},font=\footnotesize}
]
\addplot[
    color=NavyBlue,
    mark=o,
    mark size=1.5pt,
    ]
    coordinates {
    (0, 88.85)
    (.2, 91.35)
    (.4, 91.1)
    (.6, 90.65)
    (.8, 91.25)
    (1, 90.45)
    };
    \addlegendentry{\pac}
\addplot[
    color=Maroon,
    mark=triangle,
    mark size=2pt,
    ]
    coordinates {
    (0, 61.4)
    (.2, 64.85)
    (.4, 64.2)
    (.6, 63.5)
    (.8, 64.7)
    (1, 64.55)
    };
    \addlegendentry{\nac}
\addplot[
    color=Green,
    mark=diamond,
    mark size=2pt,
    ]
    coordinates {
    (0, 75.125)
    (.2, 78.1)
    (.4, 77.65)
    (.6, 77.075)
    (.8, 77.975)
    (1, 77.5)
    };
    \addlegendentry{\aac}
\end{axis}
\end{tikzpicture}

}
    \caption{Performance change for InstructGPT$_\texttt{002}$ on both tasks as the temperature changes.}
    \label{fig:temperature}
\end{figure}
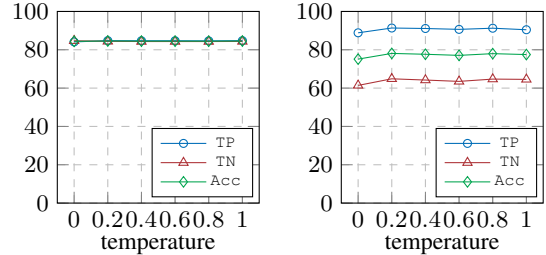

Figure~\ref{fig:temperature} shows that temperature does not influence much of the performance, thus the findings of this paper are not sensitive to temperature tuning.

\subsection{Abnormal Results of GPT-3 (\texttt{davinci})}
\label{appendix:gpt3}

\begin{figure}[t]
    \centering
	\subfigure[GPT-3 \texttt{davinci} (175B)] { 
	\label{fig:gpt3_davinci} 
		\pgfplotsset{width=0.345\linewidth,height=0.33\linewidth,compat=1.5,scale only axis}
\footnotesize
\begin{tikzpicture}
\begin{axis}[
    xlabel={$k$-shot ICL},
    xmin=-1, xmax=17,
    ymin=0, ymax=100,
    xtick={0, 2, 4, 6, 10, 16},
    legend pos=south east,
    ymajorgrids=true,
    xmajorgrids=true,
    grid style=dashed,
    x label style={at={(axis description cs:0.5,-0.125)},anchor=north},
    legend style={nodes={scale=0.7, transform shape},font=\footnotesize}
]
\addplot[
    color=NavyBlue,
    mark=o,
    mark size=1.5pt,
    error bars/.cd,
    y dir=both, y explicit
    ]
    coordinates {
    (0, 96.45)
    (2, 83.85)
    (4, 57.45)
    (6, 42.20)
    (10, 30.85)
    (16, 24.85)
    };
    \addlegendentry{\pac}
\addplot[
    color=Maroon,
    mark=triangle,
    mark size=2pt,
    error bars/.cd,
    y dir=both, y explicit
    ]
    coordinates {
    (0, 11.85)
    (2, 28.4)
    (4, 57.35)
    (6, 70.7)
    (10, 79.8)
    (16, 82.3)
    };
    \addlegendentry{\nac}
\addplot[
    color=Green,
    mark=diamond,
    mark size=2pt,
    error bars/.cd,
    y dir=both, y explicit
    ]
    coordinates {
    (0, 54.15)
    (2, 56.125)
    (4, 57.4)
    (6, 56.45)
    (10, 55.325)
    (16, 53.575)
    };
    \addlegendentry{\aac}
\end{axis}
\end{tikzpicture}

}
	\subfigure[GPT-3 \texttt{curie} (6.7B)] {
    \label{fig:gpt3_curie} 
		\pgfplotsset{width=0.345\linewidth,height=0.33\linewidth,compat=1.5,scale only axis}
\footnotesize
\begin{tikzpicture}
\begin{axis}[
    xlabel={$k$-shot ICL},
    xmin=-1, xmax=17,
    ymin=0, ymax=100,
    xtick={0, 2, 4, 6, 10, 16},
    legend pos=south east,
    ymajorgrids=true,
    xmajorgrids=true,
    grid style=dashed,
    x label style={at={(axis description cs:0.5,-0.125)},anchor=north},
    legend style={nodes={scale=0.7, transform shape},font=\footnotesize}
]
\addplot[
    color=NavyBlue,
    mark=o,
    mark size=1.5pt,
    error bars/.cd,
    y dir=both, y explicit
    ]
    coordinates {
    (0, 97.75)
    (2, 93.50)
    (4, 88.25)
    (6, 85.20)
    (10, 79.00)
    (16, 73.50)
    };
    \addlegendentry{\pac}
\addplot[
    color=Maroon,
    mark=triangle,
    mark size=2pt,
    error bars/.cd,
    y dir=both, y explicit
    ]
    coordinates {
    (0, 9.2)
    (2, 12.30)
    (4, 16.00)
    (6, 22.15)
    (10, 29.55)
    (16, 33.30)
    };
    \addlegendentry{\nac}
\addplot[
    color=Green,
    mark=diamond,
    mark size=2pt,
    error bars/.cd,
    y dir=both, y explicit
    ]
    coordinates {
    (0, 53.475)
    (2, 52.90)
    (4, 52.13)
    (6, 53.68)
    (10, 54.28)
    (16, 53.40)
    };
    \addlegendentry{\aac}
\end{axis}
\end{tikzpicture}

}
    \caption{CG results of GPT-3 for the \texttt{davinci} (175B) and \texttt{curie} (6.7B) variants, where $|E^-|=|E^+|$. Unlike other LLMs, the \nac of GPT-3 surpasses \pac when $k\ge4$.}
    \label{fig:gp3}
\end{figure}
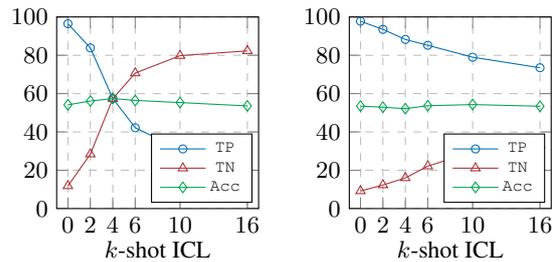

Different from the trends of other LLMs reported in $\mathsection$~\ref{sec:beliefconflict}, GPT-3 \texttt{davinci} shows a confusing pattern of the results on the CG task.
A more detailed experiment in Figure~\ref{fig:gpt3_davinci} shows that, when $k<4$, GPT-3 (\texttt{davinci}) performs similarly with its sibling LLMs, with \pac greatly surpasses \nac. 
\nac continues to enlarge as $k$ increases, even beating \pac. 
Based on \aac over the whole dataset, GPT-3 does not achieve results as good as other GPT-3 derivatives.
However, a smaller version of GPT-3 (\ie, \texttt{curie}, 6.7B) does not express such pattern, according to Figure~\ref{fig:gpt3_davinci}.
We do not have proper reasons for this finding, but further training on code and instruction tuning (\ie, Codex and InstructGPT) seem to fix this problem.

\subsection{Results of Different Relation Types}
\label{appendix:relation}

\begin{figure}[t]
    \centering
    \includegraphics[width=\linewidth]{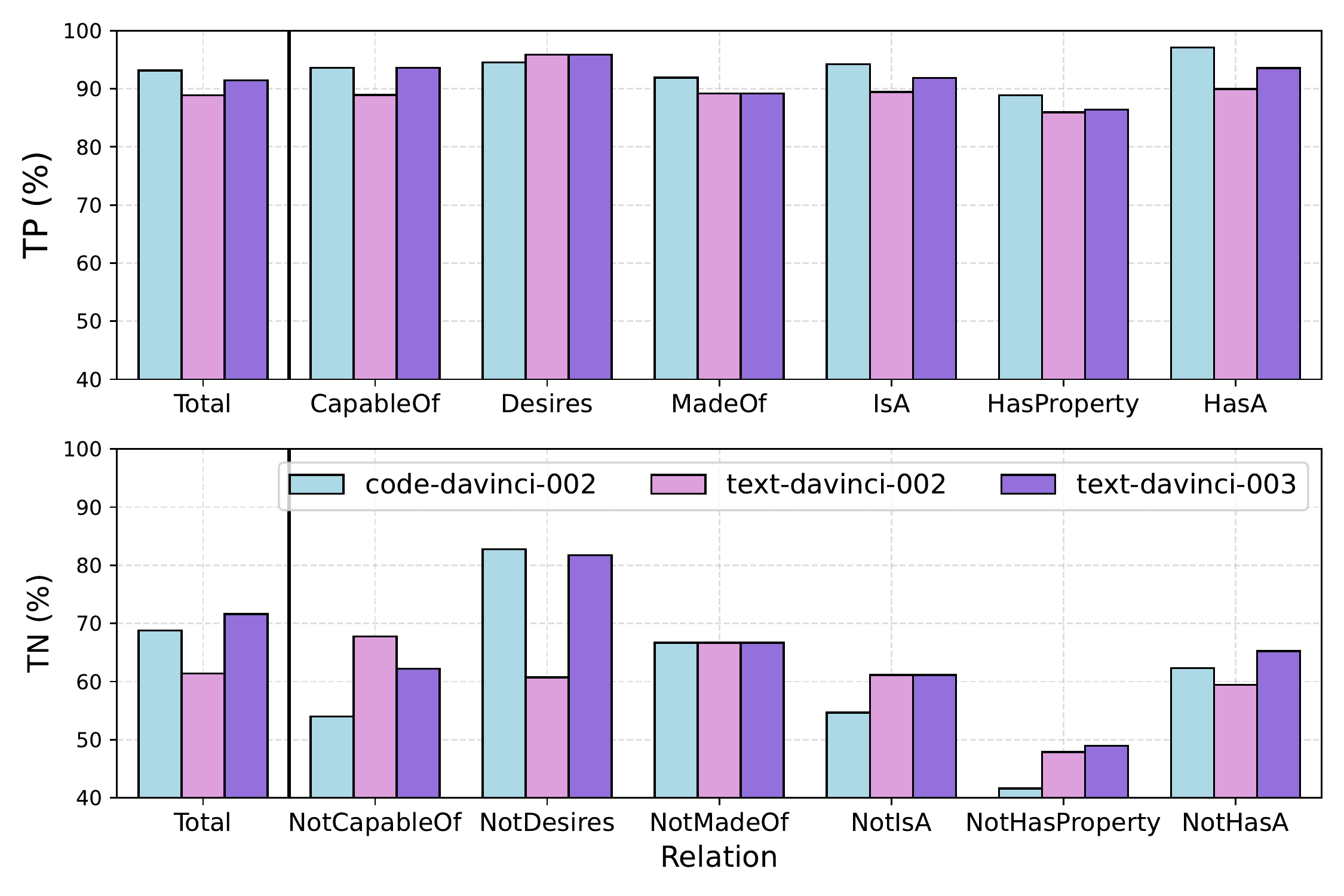}
    \caption{The 10-shot CG results per relation type on \dataset.
    The results are obtained with 10-shot learning.
    $n$ denotes the triple number per relation.
    }
    \label{fig:relation}
\end{figure}

What types of relations do LLMs find the most difficult to verbalize?
As seen in Figure~\ref{fig:relation}, we see LLMs achieve good results in the positive split.
On the negative split, LLMs unanimously believe \textsc{NotHasProperty} to be the most difficult relations.

\subsection{\textit{Do LLMs hold concerns about exceptions for commonsense knowledge?}}
\label{appendix:exception}

Commonsense knowledge usually comes with exceptions.
Could LLMs answer or generate commonsense knowledge incorrectly be because they are thinking about exceptions?
For example, ``\textit{birds can fly, but penguins cannot.}''~\cite{allaway2022penguins}.
So when asked ``\textit{can birds fly?}'', LLMs may think of a counterexample and thus arrive at the answer \textit{no}.
We rephrase the in-context examples by adding \textit{adverbs of degree} (\eg, \textit{typically}, \textit{generally}, \textit{usually}, \textit{most}, etc.) to make the tasks be about the commonsense instead of exceptions.
For instance, we rewrite ``\textit{can birds fly?}'' into ``\textit{can \underline{most} birds fly?}'' or ``\textit{can birds \underline{generally} fly?}'', and ``\textit{lions don't live in the ocean.}'' into ``\textit{lions don't \underline{usually} live in the ocean.}''
In this way, we make language explicitly convey uncertainty~\cite{reiter2019natural} and try to rule out exceptions in the tasks.

Based on the results in Table~\ref{tab:exception}, we find that adding adverbs of degree to the texts does improve LLMs' performance on both CG and QA.
This suggests that LLMs do hold a certain amount of concerns toward exceptions when dealing with commonsense reasoning, especially for negative knowledge.
However, considering exceptions with this trick still does not resolve the belief conflict.
Also, this approach could also serve as a useful trick for future commonsense research.

\begin{table}[t]
    \small
    \centering
    \begin{tabular}{cccccccc}
    \toprule
        \multirow{2}{*}{\textbf{Model}} & \multirow{2}{*}{\textbf{Exception}} & \multicolumn{3}{c}{\textbf{Perf. on QA}} & \multicolumn{3}{c}{\textbf{Perf. on CG}}  \\
        \cmidrule(lr){3-5}
        \cmidrule(lr){6-8}
        &  & \multicolumn{1}{c}{\pac} & \multicolumn{1}{c}{\nac} & \multicolumn{1}{c}{\aac} & \multicolumn{1}{c}{\pac} & \multicolumn{1}{c}{\nac} & \multicolumn{1}{c}{\aac} \\
         \midrule
        \multirow{2}{*}{\makecell[c]{Codex$_\texttt{002}$}} & - & \textbf{88.1} & \textbf{81.8} & \textbf{84.9} & \textbf{93.2} & 68.8 & 81.0 \\
        & \checkmark & 87.2 & 79.6 & 83.4 & 91.9 & \textbf{72.2} & \textbf{82.1} \\
        \cdashlinelr{1-8}
        \multirow{2}{*}{\makecell[c]{Instruct-\\GPT$_\texttt{002}$}} & - & \textbf{84.1} & 84.7 & 84.4 & 88.9 & 61.4 & 75.1 \\
        & \checkmark & 84.0 & \textbf{85.4} & \textbf{84.7} & \textbf{90.9} & \textbf{70.1} & \textbf{80.5} \\
         \bottomrule
    \end{tabular}
    \caption{10-shot QA and CG results of LLMs when adding adverbs of degree into texts, making them somehow consider exceptions of commonsense knowledge.}
    \label{tab:exception}
\end{table}

\subsection{Case Study}
\label{appendix:case_study}

\begin{table}[t]
    \centering
    \small
    \begin{tabularx}{\textwidth}{c|l|X}
        \hline
        \multicolumn{3}{c}{\textbf{Well generated negated sentence}}\\
        \hline
        1 & Triple & <deer, desires, be shoot by hunter> \\ 
        & Label & negative\\
        & Generation & Deer do \colorbox{pink}{not} desire to be shot by hunters.\\
        
        \hline
        \hline
        \multicolumn{3}{c}{\textbf{Weak negation}}\\
        \hline
         2 & Triple & <person, desires, eat alone> \\ 
        &Label&negative\\
        & Generation & \colorbox{pink}{Some} people desire to eat alone.\\
        
        \hline
        \hline
        \multicolumn{3}{c}{\textbf{Unfaithful generated sentence}}\\
        \hline
         3 & Triple & <student, desires, exam> \\ 
        &Label&negative\\
        & Generation & Students generally desire to \colorbox{pink}{do well on} exams.\\
        
        \hline
        \hline
     \multicolumn{3}{c}{\textbf{Wrong data label}}\\
        \hline
         4 & Triple & <horse, is a, bird> \\ 
        &Label&\colorbox{pink}{positive}\\
        & Generation & horses are not birds.\\
        
        \hline
        \hline
    \multicolumn{3}{c}{\textbf{High co-occurrence words}}\\
        \hline
         5 & Triple & <worm, capable of, eat bird> \\ 
        &Label& negative\\
        & Generation &  \colorbox{pink}{Worms can eat birds.}\\
        
        \hline
    \end{tabularx}
    \caption{Examples of the generated texts by InstructGPT$_{002}$ on the CG task.}
    \label{tab:case_study}
\end{table}

Table~\ref{tab:case_study} presents some examples of generated by InstructGPT$_\texttt{002}$ (10-shot).
In the 1st case, the model correctly generated negative commonsense sentences.
The 2nd one suffers from the problem of weak negation, \ie, for negative triple, the model sometimes use ``may'' or ``some'' for weak negation, which is not detected by the negation cue detector metric.
The 3rd one suffers from unfaithful generation to the constraints, where the model generates information outside the input triples to avoid generating
negation.
The 4th one is wrong due to the noise in the dataset.
The 5th one is probably due to the high co-occurrence of the concept \textit{worms} and \textit{birds}, the model finally generates a positive sentence.

\begin{table*}[t]
    \centering
    \small
    \begin{tabularx}{\textwidth}{c|l|X}
    \hline
        \multicolumn{3}{c}{\textbf{Examples for Positive Commonsense Knowledge}}\\
    \hline
        1 & Triple & <birds, capable of, fly> \\ 
        
        & Sentence & Birds can fly.\\
        
        & Question & Can birds fly? \\
        
        & Deduction & Things with lightweight bodies and strong wing muscles can usually fly. Birds have these physical characteristics. \\
        
        & Fact & Birds have wings. \\
        \hline
        2 & Triple & <playing tennis, causes, feeling relaxed>\\
        
        & Sentence & Playing tennis makes one feel relaxed.\\
        
        & Question & Does playing tennis cause someone to feel relaxed?\\
        
        & Deduction & Sport can make people feel relaxed. Tennis is a kind of sport.\\
        
        & Fact & Tennis is a kind of sport.\\
        \hline
        3 & Triple& <basketball players, desires, winning>\\
        
        & Sentence & Basketball players want to win. \\
        
        & Question & Do basketball players want to win? \\
        
        & Deduction & Winning is an important goal for many athletes. Basketball players are athletes. \\
        
        & Fact & Athletes usually desire winning in competitions.\\
        \hline
        4 & Triple & <people, desires, relax after work> \\
        
        & Sentence & People want to relax after work. \\
        
        & Question & Do people want to relaxed after work?\\
        
        & Deduction & Tired people want to relax. Work makes people tired.\\
        
        & Fact & People will be tired after work. \\
        \hline
        5 & Triple & <sheepskin, used for, writing>\\
        
        & Sentence & Sheepskin can used for writing. \\
        
        & Question & can sheepskin be used for writing? \\
        
        & Deduction & Things with a smooth and consistent surface can be used for writing. Sheepskins have that texture. \\
        
        & Fact & Sheepskin is the hide of a sheep.\\
        \hline
        \hline
        \multicolumn{3}{c}{\textbf{Examples for Negative Commonsense Knowledge}}\\
        \hline
        1 & Triple & <shoes, has a, sleeves>\\
        
        & Sentence & Shoes have no sleeve. \\
        
        & Question & Do shoes have sleeves?\\
        
        & Deduction & Sleeves are parts of garments that cover the arms. Shoes are not garments. \\
        
        & Fact & Shoe is a type of footwear.\\
        \hline
        2 & Triple & <banana, is a, tree>\\
        
        & Sentence & Bananas are not trees. \\
        
        & Question & Are bananas a kind of trees?\\
        
        & Deduction & If something is a tree, then it has an elongated trunk. Bananas do not have elongated trunks. \\
        
        & Fact & bananas are a type of fruit. \\
        \hline
        3 & Triple & <computer, is a, intelligent being>\\
        
        & Sentence & Computers aren't intelligent beings. \\
        
        & Question & Is a computer an intelligent being?\\
        
        & Deduction & Intelligent beings have the ability to think. Computers cannot think like humans do.\\
        
        & Fact &Computer is a type of electronic device.\\
        \hline
        4 & Triple & <guns, used for, healing>\\
        
        & Sentence & Guns can't be used for healing.\\
        
        & Question & Are guns used for healing?\\
        
        & Deduction & Healing instruments are tools that are used to treat injuries or illnesses. Guns are not tools that are used to treat injuries or illnesses. \\
        
        & Fact& Guns are used for killing.\\
        \hline
        5 & Triple & <elephant, capable of, jump>\\
        
        & Sentence& Elephants cannot jump.\\
        
        & Question & Can elephants jump?\\
        
        & Deduction & Jumping needs sufficient force to overcome the effects of gravity. Elephants are too heavy to overcome gravity.\\
        
        & Fact & elephants can walk slowly.\\
    \hline
    \end{tabularx}
    \caption{Some of the manually written examples used in in-context learning.}
    \label{tab:man_examples}
\end{table*}



\end{appendix}

\end{document}